\documentclass[conference]{IEEEtran}

\IEEEoverridecommandlockouts

\usepackage[dvipsnames]{xcolor}
\usepackage{amsmath,amsfonts,amssymb}
\usepackage{times}
\usepackage{graphicx}
\usepackage{multirow}
\usepackage{subcaption}
\usepackage[draft]{changes}

\usepackage{textgreek}
\usepackage{siunitx}
\usepackage[breaklinks=true]{hyperref}

\usepackage{authblk}
\usepackage{caption}

\usepackage{placeins}

\usepackage{cuted} 
\usepackage{widetext}

\usepackage[
backend=biber,
style=apa,
citestyle=authoryear-comp,
sorting=nyt,
url=false,
isbn=false,
uniquename=false,
alldates=year
]{biblatex}

\AtEveryCitekey{\clearlist{publisher}}
\AtEveryBibitem{\clearlist{publisher}}
\AtEveryBibitem{
  \clearfield{note}
}
\DeclareFieldFormat{doi}{%
    \ifhyperref
        {\href{https://doi.org/\detokenize{#1}}{\detokenize{#1}}}
        \space
}

\usepackage{rotating}
\usepackage{bbm}
\usepackage{latexsym}
\let\oldcite\cite
\renewcommand{\cite}[1]{(\oldcite{#1})}

\newcommand{\probP}{\text{I\kern-0.15em P}}
\newcommand{\had}{\circ}
\newcommand{\heavi}{H}
\renewcommand{\vec}[1]{\mathbf{#1}}
\newcommand{\mat}[1]{\mathbf{#1}}
\newcommand{\T}{^{\intercal}}

\newcommand{\appropto}{\mathrel{\vcenter{
  \offinterlineskip\halign{\hfil$##$\cr
    \propto\cr\noalign{\kern2pt}\sim\cr\noalign{\kern-2pt}}}}}

\usepackage{orcidlink}

\newcommand{\orcid}[1]{\href{https://orcid.org/#1}{\includegraphics[width=10pt]{figs/ORCIDiD_icon32x32.png}}}

\title{Vector Symbolic Finite State Machines\\in Attractor Neural Networks

\thanks{
This is the authors' final version before publication in Neural Computation.

We thank Dr. Federico Corradi, Dr. Nicoletta Risi and Dr. Matthew Cook for their invaluable input and suggestions, as well as their help with proofreading this document. 
Funded by the Deutsche Forschungsgemeinschaft (DFG, German Research Foundation) - Project MemTDE Project number 441959088 as part of the DFG priority program SPP 2262 MemrisTec - Project number 422738993, and Project NMVAC - Project number 432009531.
The authors would like to acknowledge the financial support of the CogniGron research center and the Ubbo Emmius Funds (Univ. of Groningen). 

}

}

\author[1,2,3]{Madison Cotteret \large \orcidlink{0000-0002-4891-4835}}
\author[2,3]{Hugh Greatorex \large \orcidlink{0000-0002-3716-3992}}
\author[1]{Martin Ziegler \large \orcidlink{0000-0002-6891-5747}}
\author[2,3]{Elisabetta Chicca \large \orcidlink{0000-0002-5518-8990}}

\affil[1]{\small Micro- and Nanoelectronic Systems, Institute of Micro- and Nanotechnologies (IMN) MacroNano®, \authorcr
Technische Universität Ilmenau, Ilmenau, Germany.}
\affil[2]{Bio-Inspired Circuits and Systems (BICS) Lab, Zernike Institute for Advanced Materials, University of Groningen, Netherlands.}
\affil[3]{Groningen Cognitive Systems and Materials Center (CogniGron), University of Groningen, Netherlands.}
\begin{document}

\maketitle

\thispagestyle{plain}
\pagestyle{plain}
\pagenumbering{arabic}

\begin{center} {\bf Abstract} \end{center}
% \linenumbers
Hopfield attractor networks are robust distributed models of human memory, but lack a general mechanism for effecting state-dependent attractor transitions in response to input. We propose construction rules such that an attractor network may implement an arbitrary finite state machine (FSM), where states and stimuli are represented by high-dimensional random vectors, and all state transitions are enacted by the attractor network's dynamics. Numerical simulations show the capacity of the model, in terms of the maximum size of implementable FSM, to be linear in the size of the attractor network for dense bipolar state vectors, and approximately quadratic for sparse binary state vectors. We show that the model is robust to imprecise and noisy weights, and so a prime candidate for implementation with high-density but unreliable devices. By endowing attractor networks with the ability to emulate arbitrary FSMs, we propose a plausible path by which FSMs could exist as a distributed computational primitive in biological neural networks.
%%%%%%%%%%%

\vspace{1em}


\section{Introduction}

Hopfield attractor networks are one of the most celebrated models of robust neural auto-associative memory, as from a simple Hebbian learning rule they display emergent attractor dynamics which allow for reliable pattern recall, completion, and correction even in situations with considerable non-idealities imposed \cite{hopfield_neural_1982, amit_modeling_1989}. Attractor models have since found widespread use in neuroscience as a functional and tractable model of human memory \cite{little_existence_1974, schneidman_weak_2006, chaudhuri_computational_2016, khona_attractor_2022, rolls_mechanisms_2013, eliasmith_unified_2005}. The assumption of these models is that the network represents different states by different, usually uncorrelated, global patterns of persistent activity. When the network is presented with an input that closely resembles one of the stored states, the network state converges to the corresponding fixed-point attractor.

This process of switching between discrete attractor states is thought to be fundamental both to describe biological neural activity, as well as to model higher cognitive decision making processes \cite{daelli_neural_2010, mante_context-dependent_2013, miller_itinerancy_2016, tajima_task-dependent_2017, brinkman_metastable_2022}. What attractor models currently lack, however, is the ability to perform state-dependent computation, a hallmark of human cognition \cite{dayan_simple_2008, buonomano_state-dependent_2009, granger_toward_2020}. That is, when the network is presented with an input, the attractor state to which the network switches ought to be dependent both upon the input stimulus as well as the state the network currently inhabits, rather than simply the input.

We thus seek to endow a classical neural attractor model, the Hopfield network, with the ability to perform state-dependent switching between attractor states, without resorting to the use of biologically implausible mechanisms, such as training via backpropagation algorithms. The resulting attractor networks will then be able to robustly emulate any arbitrary Finite State Machine (FSM), considerably improving their usefulness as a neural computational primitive.

We achieve this by leaning heavily on the framework of Vector Symbolic Architectures (VSAs), also known as Hyperdimensional Computing (HDC). VSAs treat computation in an entirely distributed manner, by letting symbols be represented by high-dimensional random vectors, hypervectors \cite{kanerva_fully_1997, plate_holographic_1995, gayler_multiplicative_1998, kleyko_survey_2022}. When equipped with a few basic operators for binding and superimposing hypervectors together, corresponding often either to component-wise multiplication or addition respectively, these architectures are able to store primitives such a sets, sequences, graphs and arbitrary data bindings, as well as enabling more complex relations, such as analogical and figurative reasoning \cite{kanerva_hyperdimensional_2009, kleyko_vector_2021}. Although different VSA models often have differing representations and binding operations \cite{kleyko_survey_2022}, they all share the need for an auto-associative cleanup memory, which can recover a clean version of the most similar stored hypervector, given a noisy version of itself. We here use the recurrent dynamics of a Hopfield-like attractor neural network as a state-holding auto-associative memory \cite{gritsenko_neural_2017}.

Symbolic FSM states will thus be represented each by a hypervector and stored within the attractor network as a fixed-point attractor. Stimuli will also be represented by hypervectors, which, when input to the attractor network, will trigger the network dynamics to transition between the correct attractor states. We make use of common VSA techniques to construct a weights matrix to achieve these dynamics, where we use the Hadamard product between bipolar hypervectors $\{-1,1\}^N$ as the binding operation (the Multiply-Add-Permute (MAP) VSA model) \cite{gayler_multiplicative_1998}. We thus claim that attractor-based FSMs are a plausible biological computational primitive insofar as Hopfield networks are.

This represents a computational paradigm that is a departure from conventional von Neumann architectures, wherein the separation of memory and computation is a major limiting factor in current advances in conventional computational performance (the von Neumann bottleneck \cite{backus_can_1978, indiveri_memory_2015}). Similarly, the high redundancy and lack of reliance on individual components makes this architecture fit for implementation with novel in-memory computing technologies such as resistive RAM (RRAM) or phase-change memory (PCM) devices, which could perform the network's matrix-vector-multiplication (MVM) step in a single operation \cite{xia_memristive_2019, ielmini_-memory_2018, zidan_chapter_2020}.

\section{Methods}

\subsection{Hypervector arithmetic}

Throughout this article, symbols will be represented by high-dimensional randomly-generated dense bipolar hypervectors
\begin{equation}
    \vec{x} \in \{ -1 , 1\}^{N}
\end{equation}
where the number of dimensions $N$ is generally taken to be greater than $10,000$. Unless explicitly stated otherwise, any bold lowercase Latin letter may be assumed to be a new, independently generated hypervector, with the value $Y_{i}$ at any index $i$ in $\vec{x}$ generated according to
\begin{equation}
    \probP(Y_{i} = 1) = \probP(Y_{i}=-1) = \frac{1}{2}
\label{eqn:probX_half}
\end{equation}
For any two arbitrary hypervectors $\vec{a}$ and $\vec{b}$, we define the similarity between the two hypervectors by the normalised inner product
\begin{equation}
    d(\vec{a},\vec{b}) := \frac{1}{N} \vec{a} \T \vec{b} = \frac{1}{N} \sum_{i = 1}^{N} a_{i}b_{i}
\label{eqn:d_simple}
\end{equation}
where the similarity between a hypervector and itself $d(\vec{a},\vec{a}) = 1$, and $d(\vec{a},-\vec{a}) = -1$. Due to the high dimensionality of the hypervectors, the similarity between any two unrelated (and so independently generated) hypervectors is the mean of an unbiased random sequence of $-1$ and $1$s
\begin{equation}
    d(\vec{a}, \vec{b}) = 0 \pm \frac{1}{\sqrt{N}} \approx 0
\end{equation}
which tends to 0 for $N\rightarrow \infty$. It is from this result that we get the requirement of high dimensionality, as it ensures that the inner product between two random hypervectors is approximately 0. We can thus say that independently generated hypervectors are \textit{pseudo-orthogonal} \cite{kleyko_vector_2021}. For a set of independently generated states $\{\vec{x}^\mu\}$, these results can be summarised by
\begin{equation}
    d(\vec{x}^{\mu},\pm \vec{x}^{\nu} ) \, \stackrel{N \rightarrow \infty}{=}  \,  \pm \delta^{\mu \nu}
\end{equation}
where $\delta^{\mu \nu}$ is the Kronecker delta. Hypervectors may be combined via a so called \textit{binding} operation to produce a new hypervector that is dissimilar to both its constituents. We here choose the Hadamard product, or component-wise multiplication, as our binding operation, denoted ``$\had$''.
\begin{equation}
    (\vec{a}\had \vec{b})_{i} = a_{i} \cdot b_{i}
\end{equation}

The statement that the binding of two hypervectors is dissimilar to its constituents is written as
\begin{equation}
\begin{split}
    d(\vec{a} \had \vec{b}, \vec{a}) \approx 0 \\
    d(\vec{a} \had \vec{b}, \vec{b}) \approx 0
\end{split}
\end{equation}
where we implicitly assume that $N$ is large enough that we can ignore the $\mathcal{O}(\frac{1}{\sqrt{N}})$ noise terms.
If we wish to recover a similarity between the hypervectors $\vec{a} \had \vec{b}$ and $\vec{a}$, we could bind the $\vec{b}$ hypervector to the lone $\vec{a}$ term as well, in which case we would have $d(\vec{a} \had \vec{b}, \vec{a} \had \vec{b}) = 1$. For reasons of ease and robustness of implementation in an asynchronous neural system, we focus instead on another method to recover the similarity (see Sections \ref{section:W_construction} \& \ref{subsec:hadamard_input}). If we \textit{mask} the system using $\vec{b}$, such that only components where $b_i = 1$ are remaining. Then, we have
\begin{equation}
\begin{split}
d \big( \vec{a} \had \vec{b} &, \vec{a} \had \heavi(\vec{b}) \big)  = \frac{1}{N} \sum_{1 \leq i \leq N} a_i b_i a_i H(b_i) \\
& = \frac{1}{N} \Bigg[ \sum_{\substack{1 \leq i \leq N \\ b_i =1}} a_i^2 H(1) -  \! \! \sum_{\substack{1 \leq i \leq N \\ b_i =-1}} \! \! a_i^2 H(-1) \Bigg]  \\
& = \frac{1}{N} \sum_{\substack{1 \leq i \leq N \\ b_i =1}} 1 \\
& \approx \frac{1}{2} 
\end{split}
\end{equation}
where we have used the Heaviside step function $\heavi(\cdot)$ defined by
\begin{equation}
    \big( \heavi(\vec{b}) \big)_{i} = \heavi(b_{i}) = \begin{cases} 1 \quad \text{if} \quad b_{i} > 0 \\ 0 \quad \text{otherwise} \end{cases}
\end{equation}
to create a multiplicative mask $\heavi(\vec{b})$, setting to 0 all components where $b_i = -1$. In the second line, we have split the summation over all components into summations over components where $b_i = 1$ and $-1$ respectively. The final similarity of $\frac{1}{2}$ is a consequence of approximately half of all values in a any hypervector being $+1$ (\mbox{Equation \ref{eqn:probX_half}}).

\subsection{Hopfield networks}

A Hopfield network is a dynamical system defined by its internal state vector $\vec{z}$ and fixed recurrent weights matrix $\mat{W}$, with a state update rule given by
\begin{equation}
    \vec{z}_{t+1} = \mathrm{sgn} \big( \mat{W} \vec{z}_t \big)
    \label{eqn:update_hopfield}
\end{equation}
where $\vec{z}_{t}$ is the network state at discrete time step $t$, and $\mathrm{sgn}(\cdot)$ is an component-wise sign function, with zeroes resolving\footnote{Though this arbitrary choice may seem to incur a bias to a particular state, in practise the postsynaptic sum very rarely equals 0.} to +1 \cite{hopfield_neural_1982}.
We know that if we want to store $P$ uncorrelated patterns $\{ \vec{x}^{\nu} \}_{\nu =1}^{P}$ within a Hopfield network, we can construct the weights matrix $\mat{W}$ according to
\begin{equation}
    \mat{W} = \sum_{\mathrm{patterns} \, \, \nu}^{P} \vec{x}^{\nu} \vec{x}^{\nu \intercal}
\end{equation}
then as long as not too many patterns are stored ($P < 0.14 N$ \cite{hopfield_neural_1982}), the patterns will become fixed-point attractors of the network's dynamics, and the network can perform robust auto-associative pattern completion and correction.

% Since we are operating in high dimensional space, this condition is more than satisfied.

\subsection{Finite State Machines}

A Finite State Machine (FSM) $M$ is a discrete system with a finite state set $X_{\text{FSM}} = \{\chi_{1},\chi_{2}, \ldots , \chi_{N_{Z}}$ \}, a finite input stimulus set $S_{\text{FSM}} = \{\varsigma_{1}, \varsigma_{2}, \ldots, \varsigma_{N_{S}} \}$, and finite output response set $R_{\text{FSM}} = \{\rho_{1}, \rho_{2}, \ldots, \rho_{N_{R}} \}$. The FSM $M$ is then fully defined with the addition of the transition function $F(\cdot) : X_{\text{FSM}} \times S_{\text{FSM}} \rightarrow X_{\text{FSM}}$ and the output response function $G(\cdot) : X_{\text{FSM}} \times S_{\text{FSM}} \rightarrow R_{\text{FSM}}$
\begin{equation}
\begin{split}
    x_{t+1} & = F(x_{t}, s_{t}) \\
    r_{t+1} & = G(x_{t}, s_{t})
\end{split}
\end{equation}
where $x_{t} \in X_{\text{FSM}}$, $r_{t} \in R_{\text{FSM}}$ and $s_{t} \in S_{\text{FSM}}$ are the state, output and stimulus at time step $t$ respectively. The transition function $F(\cdot)$ thus provides the next state for any state-stimulus pair, while $G(\cdot)$ provides the output, and both may be chosen arbitrarily. The FSM $M$ can thus be represented by a directed graph, where each node represents a different state $\chi$, and every edge has a stimulus $\varsigma$ and optional output $\rho$ associated with it.

\section{Attractor network construction}
\label{section:methods}

We now show how a Hopfield-like attractor network may be constructed to emulate an arbitrary FSM, where the states within the FSM are stored as attractors in the network, and the stimuli for transitions between FSM states trigger all corresponding transitions between attractors. More specifically, for every FSM state $\chi \in X_{\text{FSM}}$, an associated hypervector $\vec{x}$ is randomly generated and stored as an attractor within the network, the set of which we denote $X_{\text{AN}}$. We henceforth refer to these hypervectors as node hypervectors, or node attractors. 
% We use $X_{\text{AN}}$ to denote the set of node hypervectors stored as attractors within the network.
Every unique stimulus $\varsigma \in S_{\text{FSM}}$ in the FSM is also now associated with a randomly generated hypervector $\vec{s} \in S_{\text{AN}}$, where $S_{\text{AN}}$ is the set of all stimulus hypervectors. For the FSM edge outputs $\rho \in R_{\mathrm{FSM}}$, a corresponding set of output hypervectors $\vec{r} \in R_\mathrm{AN}$ is similarly generated. These correspondences are summarised in \mbox{Table \ref{table:notation}}.

\begin{table}[h]
\centering
\begin{tabular}{ll|ll}
\multicolumn{2}{c|}{FSM (Symbols)}    & \multicolumn{2}{c}{Attractor Network (Hypervectors)} \\ \hline
States                      &  $\chi \in X_{\text{FSM}}$  & Attractors           & $\vec{x} \in X_{\text{AN}}$             \\
Stimuli                      & $\varsigma \in S_{\text{FSM}}$ & Stimuli            & $\vec{s} \in S_{\text{AN}}$             \\
Outputs                      & $\rho \in R_{\text{FSM}}$  & Outputs            & $\vec{r} \in R_{\text{AN}}$             
% 7                      & 3  & 8            & f            
\end{tabular}
\caption{A comparison of the notation used to represent states, stimuli and outputs in the FSM, and the corresponding hypervectors used to represent the FSM within the attractor network.}
\label{table:notation}
\end{table}

\subsection{Constructing transitions}
\label{section:W_construction}

We consider the general situation that we want to initiate a transition from \textit{source} attractor state $\vec{x} \in X_{\text{AN}}$ to \textit{target} attractor state $\vec{y} \in X_{\text{AN}}$, by imposing some stimulus hypervector $\vec{s} \in S_{\text{AN}}$ as input onto the network.
\begin{equation}
    \vec{x} \stackrel{\vec{s}}{\longrightarrow} \vec{y}
\end{equation}

To ensure the plausible functionality of the network in a biological system, the mechanism for enacting transitions in the network should make very few timing assumptions about the system, and should be robust to an arbitrary degree of asynchrony. How we model input to the network is thus of crucial importance to its functionality in these regimes. We model input to the network as a \textit{masking} of the network state, such that all components where the stimulus $\vec{s}$ is -1 are set to 0. This may be likened to saying we are considering input to the network that selectively silences half of all neurons according to the stimulus hypervector. This mechanism was chosen as it allows the network to function even when the input is applied asynchronously and with random delays (see Section \ref{subsec:hadamard_input}). While a stimulus hypervector $\vec{s}$ is being imposed upon the network, the modified state update rule is given by
\begin{equation}
    \vec{z}_{t+1} = \mathrm{sgn} \big( \mat{W} (\vec{z}_{t} \had \heavi (\vec{s})) \big)
\label{eqn:update_hop_mask}
\end{equation}
where the Hadamard product of the network state with $\heavi (\vec{s})$ enacts the masking operation, and the weights matrix $\mat{W}$ is constructed such that $\vec{z}_{t+1}$ will resemble the desired target state (Section \ref{section:W_construction}).

For every edge in the FSM, we randomly generate an ``edge state'' $\vec{e}$, which is also stored as an attractor within the network. Each edge will use this $\vec{e}$ state as an intermediate attractor state, en route to $\vec{y}$. Additionally, each unique stimulus $\varsigma \in S_\mathrm{FSM}$ will now have \textit{two} stimulus hypervectors associated with it, $\vec{s}_a$ and $\vec{s}_b$, which trigger transitions from source state $\vec{x}$ to edge state $\vec{e}$ and edge state $\vec{e}$ to target state $\vec{y}$ respectively. The edge states are introduced to allow the system to function even when stimuli are input to the network for arbitrarily many time steps, and prevents unwanted effects such as skipping over certain attractor states, or oscillations between states (see Section \ref{sec:why_edge_states}). A general transition now looks like
\begin{equation}
    \vec{x} \stackrel{\vec{s}_a}{\longrightarrow} \vec{e} \stackrel{\vec{s}_b}{\longrightarrow} \vec{y}
\end{equation}
where $\vec{x}, \vec{y} \in X_{\text{AN}}$ are node attractor states but $\vec{e}$ exists purely to facilitate the transition.
The weights matrix is constructed\footnote{We have here ignored that the diagonal of $\mat{W}$ is set to 0 (no self connections), but this does not significantly affect the following results.} as
\begin{equation}
    \mat{W} = \underbrace{\frac{1}{N}\sum_{\vphantom{\text{dg}}\text{nodes $\nu$}}^{N_{Z}} \vec{x}^\nu \vec{x}^{\nu \intercal} }_{\substack{\text{ Hopfield}\\\text{attractor terms}}} +
    \underbrace{\frac{1}{N}\sum_{\text{edges $\eta$}}^{N_{E}} \mat{E}^{\eta}}_{\substack{\text{Asymmetric}\\\text{transition terms}}}
\label{eqn:W_construction}
\end{equation}
where $\vec{x}^\nu \in X_{\text{AN}}$ is the node hypervector corresponding to the $\nu$'th node in the graph to be implemented, $N_{Z}$ and $N_{E}$ are the number of nodes and edges respectively, and $\mat{E}^\eta$ is the addition to the weights matrix required to implement an individual edge, given by
\begin{equation}
\begin{split}
    \mat{E}^{(\eta)} & = \vphantom{(}\vec{e} \vec{e}\T  \\
    & +\heavi(\vec{s}_a) \had (\vec{e}-\vec{x}) (\vec{x}  \had  \vec{s}_a)\T \\
    & +\heavi(\vec{s}_b) \had (\vec{y} -\vec{e}) (\vec{ e}  \had  \vec{s}_b)\T 
    \label{eqn:tran_term}
\end{split}
\end{equation}

where $\vec{x}$, $\vec{e}$ and $\vec{y}$ are the source, edge, and target states of the edge $\eta$ respectively, and $\vec{s}_a$ and $\vec{s}_b$ are the input stimulus hypervectors associated with this edge's label. The edge index $\eta$ has been dropped for brevity. The $\vec{e}\vec{e}\T$ term is the edge state attractor we have introduced as an intermediary for the transition. The second set of terms enacts the $ \vec{x} \stackrel{\vec{s}_a}{\longrightarrow} \vec{e}$ transition, by giving a nonzero inner product with the network state $\vec{z}_t$ only when the network is in state $\vec{x}$, \textit{and} the network is being masked by the stimulus $\vec{s}_a$. When both of these conditions are met, the $(\vec{x} \had \vec{s}_a)^\intercal$ term will have a nonzero inner product with the network state, projecting out the $(\vec{e} - \vec{x})$ term, which ``pushes'' the network from the $\vec{x}$ to the $\vec{e}$ attractor state. This allows terms to be stored in $\mat{W}$ which are effectively obfuscated, not affecting network dynamics considerably, until a specific stimulus is applied as a mask to the network. Likewise, the third set of terms enacts the $\vec{e} \stackrel{\vec{s}_b}{\longrightarrow} \vec{y}$ transition.

In the absence of input, the network functions like a standard Hopfield attractor network,
\begin{equation}
    \mat{W}\vec{x} \approx \vec{x} \pm \sigma \vec{n} \quad \forall \quad \vec{x} \in X_{\text{AN}}
\label{eqn:normal_attractor}
\end{equation}
where $\vec{n} \in \mathbb{R}^{N}$ is a standard normally-distributed random vector, and
\begin{equation}
\sigma = \sqrt{\frac{N_{Z} + 3N_{E}}{N}}   
\label{eqn:sigma_SNR}
\end{equation} 
is the magnitude of noise due to the undesired finite inner product with other stored terms (see Section \ref{section:appendix_no_mask} for proof). Thus as long as the magnitude of the noise is not too large, $\vec{x}$ will be a solution of $\vec{z} = \mathrm{sgn}(\mat W\vec{z})$ and so a fixed-point attractor of the dynamics.

When a valid stimulus is presented as input to the network however, masking the network state, the previously obfuscated asymmetric transition terms become significant and dominate the dynamics. Assuming there is a stored transition term $\mat{E}$ corresponding to a valid edge with hypervectors $\vec{x},\vec{e},\vec{y},\vec{s}_a, \vec{s}_b$ having the same meaning as in \mbox{Equation \ref{eqn:tran_term}}, during a masking operation we have 
\begin{equation}
\begin{split}
    \mat{W} \big( \vec{x} \had \heavi(\vec{s}_a) \big)   \appropto & \, \,  \underbrace{\vphantom{dg}\heavi \big( \vec{s}_a \big) \had \vec{e}}_{\substack{\text{Projection to} \\ \text{edge state}}} \, + \, \underbrace{\vphantom{dg}\heavi(-\vec{s}_a)\had \vec{x}}_{\substack{\text{May be} \\ \text{ignored}}} \,\pm \sqrt{2}\sigma \vec{n}
\end{split}
\end{equation}
where $\appropto$ implies approximate proportionality (see Section \ref{section:appendix_mask} for proof). The second set of terms can be ignored, as they project only to neurons which are currently being masked. Thus the only significant term is that containing the edge state $\vec{e}$, which consequently drives the network to the $\vec{e}$ state, enacting the $\vec{x} \stackrel{\vec{s}_a}{\longrightarrow} \vec{e}$ transition. Since the state $\vec{e}$ is also stored as an attractor within the network, we have
\begin{equation}
    \mat{W} \big( \vec{e} \had \heavi(\vec{s}_a) \big) \, \appropto \,  \vec{e} \pm \sqrt{2} \sigma \vec{n}   
\end{equation}
and
\begin{equation}
    \mat{W}\vec{e} \approx \vec{e} \pm \sigma \vec{n}
\end{equation}
thus the edge states $\vec{e}$ are also fixed-point attractors of the network dynamics. To complete the transition from state $\vec{x}$ to $\vec{y}$, the second stimulus $\vec{s}_b$ is applied, giving
\begin{equation}
\begin{split}
    \mat{W} \big( \vec{e} \had \heavi(\vec{s}_b) \big)  & \appropto  \heavi \big( \vec{s}_b \big) \had \vec{y} + \heavi(-\vec{s}_b) \had \vec{e} \pm \sqrt{2}\sigma \vec{n}
\end{split}
\end{equation}
which drives the network state towards $\vec{y} \in X_{\text{AN}}$, the desired target attractor state. By consecutive application of the inputs $\vec{s}_a$ and $\vec{s}_b$, the transition terms $\mat{E}^\eta$ stored in $\mat{W}$ have thus caused the network to controllably transition from the source state attractor state to the target attractor state. Due to the robustness of the masking mechanism, the stimuli can be applied asynchronously and with arbitrary delays (see Section \ref{subsec:hadamard_input}). Transition terms $\mat{E}^\eta$ may be iteratively added to $\mat{W}$ to achieve any arbitrary transition between attractor states, and so any arbitrary FSM may be implemented within a large enough attractor network.

\subsection{Edge outputs}

Until now we have not mentioned the other critical component of FSMs: the output associated with every edge. We have separated the construction of transitions and edge outputs for clarity, since the two may be effectively decoupled.
Much like for the nodes and edges in the FSM to be implemented, for every unique FSM output $\rho \in R_{\text{FSM}}$, we generate a corresponding hypervector $\vec{r} \in R_{\text{AN}}$, where $R_{\text{AN}}$ is the set of all output hypervectors.
We then seek to somehow embed these hypervectors into the attractor network, such that every transition between node attractor states may contain one of these hypervectors $\vec{r}$. A natural solution would be to embed the $\vec{r}$ hypervector into the edge state attractors $\vec{e}\vec{e}\T$, since there already exists one for every edge. We can consider altering the edge state attractors from $\vec{e}\vec{e}\T$ to ${\vec{e}_\vec{r} \vec{e}_\vec{r}}\T$, where $\vec{e}_\vec{r}$ resembles the original $\vec{e}$ state with $\vec{r}$ somehow embedded within it, such that its presence can be detected via a linear projection. If multiple edges have the same $\vec{r}$ hypervector however, then the ${\vec{e}_\vec{r} \vec{e}_\vec{r}}\T$ terms for different edges will be correlated, incurring unwanted interference between attractor states and violating the assumption that the inner product between different attractor terms is small enough that it can be ignored. We avoid this by instead storing altered edge state attractors of the form $\vec{e}_\vec{r} \vec{e}\T$. We then choose $\vec{e}_\vec{r}$ such that it is minimally different from $\vec{e}$ (i.e. $d(\vec{e}_\vec{r}, \vec{e}) \approx 1$), so that we still retain the desired attractor dynamics. We thus choose the output hypervectors $\vec{r} \in R_{\mathrm{AN}}$ to be sparse ternary hypervectors $\vec{r} \in \{ -1, 0, 1 \}^{N}$ with coding level $f_r := \frac{1}{N}\sum_{i}^N \lvert r_i \rvert $, the fraction of nonzero components. These output hypervectors are then embedded in the edge state attractors, altering the $\vec{e}\vec{e}\T$ terms in each $\mat{E}$ term according to
\begin{equation}
    \vec{e}\vec{e}\T \rightarrow \vec{e}_{\vec{r}}\vec{e}\T := \Big[ \vec{e}  \had  \big( \vec{1}-\heavi(\vec{r} \had \vec{r}) \big) +  \vec{r} \Big] \vec{e} \T 
\label{eqn:output_attr}
\end{equation}
where the composite vector $\vec{e}_{\vec{r}}$ introduced above is here defined and $\vec{1}$ is a hypervector of all ones. As a result of this modification, the edge states $\vec{e}$ themselves will no longer be exact attractors of the space. The composite state $\vec{e}_{\vec{r}}$ will however be stable, in which the presence of $\vec{r}$ can be easily detected by a linear projection ($\vec{e}_{\vec{r}} \cdot \vec{r} = Nf_{r}$). This has been achieved without incurring any similarity and thus interference between attractors, which would otherwise alter the dynamics of the previously described transitions.
A full transition term $\mat{E}^\eta$, including its output, is thus given by
\begin{equation}
\begin{split}
    \mat{E}^{(\eta)} & = \vphantom{(}\Big[ \vec{e}  \had  \big( \vec{1}-\heavi(\vec{r} \had \vec{r}) \big) +  \vec{r} \Big] \vec{e} \T  \\
    & +\heavi(\vec{s}_a) \had (\vec{e}-\vec{x}) (\vec{x}  \had  \vec{s}_a)\T \\
    & +\heavi(\vec{s}_b) \had (\vec{y} -\vec{e}) (\vec{ e}  \had  \vec{s}_b)\T 
    \label{eqn:tran_term_final}
\end{split}
\end{equation}
which combined with the network state masking operation is solely responsible for storing the FSM connectivity and enabling the desired inter-attractor transition dynamics.

\subsection{Sparse activity states}
\label{section:sparse_theory}

It is well known that the memory capacity of attractor networks can be vastly increased by storing sparsely-coded activity patterns, rather than dense patterns as we have done thus far \cite{amari_characteristics_1989, tsodyks_enhanced_1988, amit_modeling_1989}. We therefore adapt the construction of the attractor network to the case that the network state $\vec{z}_t$ and its stored hypervectors $\vec{x}^{\nu}$ are binary and $f-$sparse, i.e. contain mostly zeroes, with very few entries being $+1$, to test if there are similar gains in the size of FSM that can be reliably embedded. To distinguish these hypervectors from the dense bipolar hypervectors we have been using thus far, we denote sparse binary hypervectors $\vec{x}_{\mathrm{sp}} \in \{0,1\}^{N}$ with $|\vec{x}_{\mathrm{sp}}|_1 = N f$, where $f$ is the fixed coding level of the states, the fraction of nonzero components. Note that we here construct hypervectors which have \textit{exactly} $Nf$ nonzero components, and so they may better be described as a sparse $N$-of-$M$ code \cite{furber_sparse_2004}. The attractor network's weights matrix is constructed as
\begin{equation}
    \mat{W} = \sum_\nu (\vec{x}^\nu_{\mathrm{sp}} - f \vec{1}) (\vec{x}^\nu_{\mathrm{sp}} - f \vec{1})^\intercal + \sum_\eta \mat{E}^{\eta}
\end{equation}
where $\mat{E}^{\eta}$ are the equivalent sparse edge terms to be defined. If the neuron state update rule \mbox{(Equation \ref{eqn:update_hopfield})} is replaced with a sparse binary variant, e.g. a top-$k$ activation function or a Heaviside function with an appropriately chosen threshold, then the stored states $\vec{x}^{\nu}_{\mathrm{sp}}$ will be attractors of the network's dynamics \cite{amari_characteristics_1989}. The additional edge terms $\mat{E}^{\eta}$ are analogously constructed as
\begin{equation}
\begin{split}
    \mat{E}^{(\eta)} & = \vphantom{(}(\vec{e}_{\mathrm{sp}}-f\vec{1}) (\vec{e}_{\mathrm{sp}} - f\vec{1})\T  \\
    & + (\vec{e}_{\mathrm{sp}}-\vec{x}_{\mathrm{sp}}) \big( (\vec{x}_{\mathrm{sp}}-f \vec{1})  \had  \vec{s}_a \big) \T \\
    & + (\vec{y}_{\mathrm{sp}} -\vec{e}_{\mathrm{sp}}) \big( (\vec{ e}_{\mathrm{sp}} -f \vec{1})  \had  \vec{s}_b \big)\T 
    \label{eqn:tran_term_sparse}
\end{split}
\end{equation}
where the first set of terms embeds the sparse binary edge state $\vec{e}_{\mathrm{sp}}$ as an attractor, while the second and third terms embed the source-to-edge and edge-to-target transitions respectively.
% Note that the stimulus terms $\vec{s}_a$ and $\vec{s}_b$ remain dense bipolar hypervectors, to ensure that the bound terms yield negligible overlap with the network state when the stimulus is not being imposed as a mask.
The stimulus hypervectors $\vec{s}_a$ and $\vec{s}_b$ can also be made sparse, such that fewer than half of all neurons are masked by the stimuli, but at the cost of decreased memory capacity (Section \ref{section:sparse_stimuli}). For this reason, we here keep them as bipolar hypervectors, with an approximately equal number of $+1$ as $-1$ entries.  Each set of terms within each $\mat{E}^{\eta}$ term performs the same role as in the dense bipolar case as discussed in Section \ref{section:W_construction}. How output states should be embedded into each transition in the sparse case is unclear, because unlike in the dense case, they cannot be embedded into the edge state attractors without considerably affecting the network dynamics and thus attractor stabilities.

\section{Results}

\subsection{FSM emulation}

\begin{figure}
\centering
\includegraphics[width=3in]{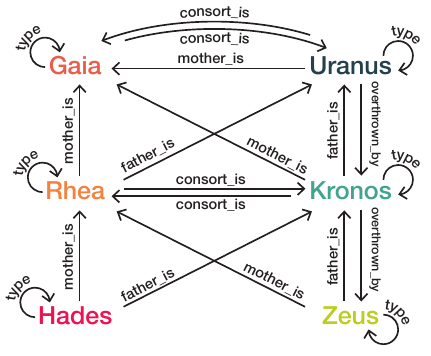}
\caption{An example FSM which we implement within the attractor network. Each node within the graph (e.g. ``Zeus'') is represented by a new hypervector $\vec{x}^\mu$ and stored as an attractor within the network. Every edge is labelled by its stimulus (e.g. ``father\_is''), for which corresponding hypervectors $\vec{s}_a$ and $\vec{s}_b$ are also generated. When a stimulus' hypervector is input to the network, it should allow all corresponding attractor transitions to take place. Each edge may also have an associated output symbol, where we here choose the edges labelled ``type'' to output the generation of the god \{``Primordial'', ``Titans'', ``Olympians''\}. This graph was chosen as it displays the generality of the embedding: it contains cycles, loops, bidirectional edges and state-dependent transitions.
}
\label{fig:starwars_net}
\end{figure}

\begin{figure*}
\centering
\includegraphics[width=0.9\linewidth]{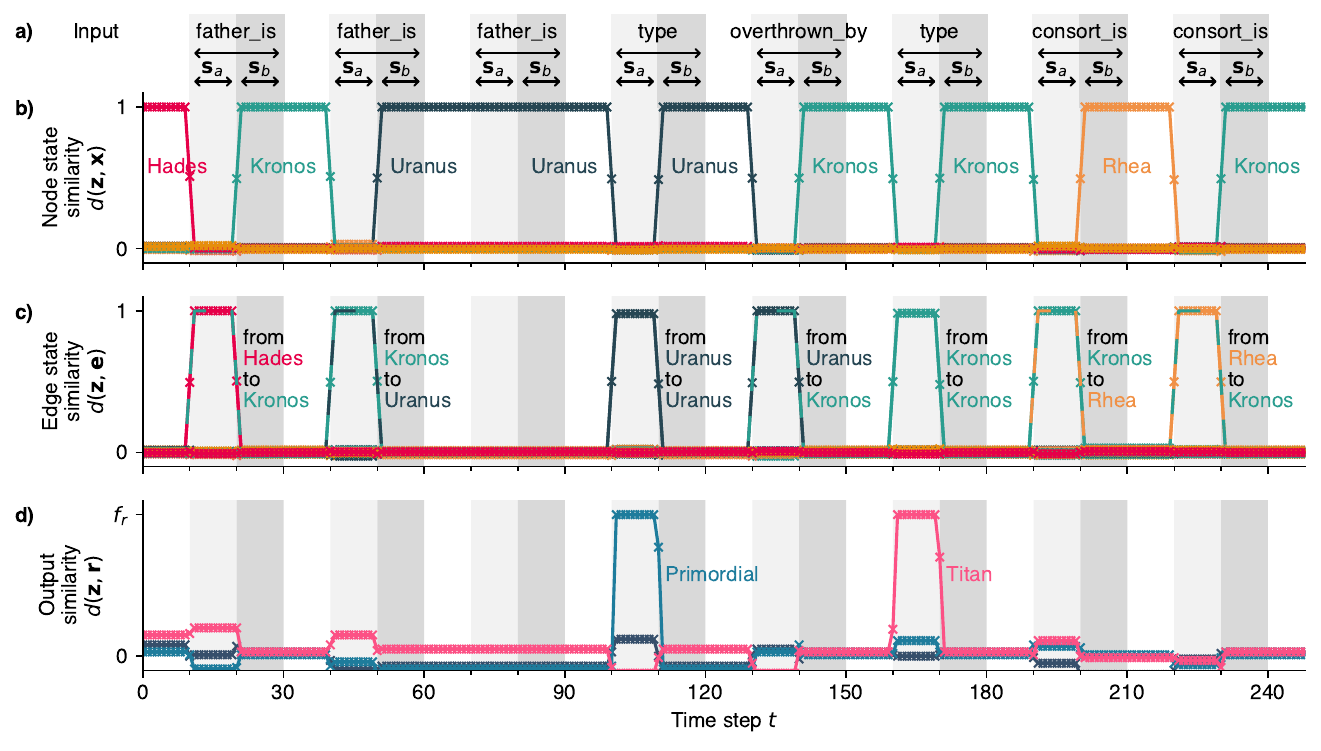}
\caption{An attractor network transitioning through attractor states in a state-dependent manner, as a sequence of input stimuli is presented to the network. \textbf{a)} The input stimuli to the network, where for each unique stimulus (e.g. ``father\_is'' in the FSM to be implemented (\mbox{Figure \ref{fig:starwars_net}}) a pair of hypervectors $\vec{s}_a$ and $\vec{s}_b$ have been generated. No stimulus, a stimulus $\vec{s}_a$, then a stimulus $\vec{s}_b$ are input for 10 time steps each in sequence. \textbf{b)} \& \textbf{c)} The similarity of the network state $\vec{z}_t$ to stored node attractor states $\vec{x} \in X_{\text{AN}}$ and stored edge states $\vec{e}$ respectively, computed via the inner product (\mbox{Equation \ref{eqn:d_simple}}). \textbf{d)} The similarity of the network state $\vec{z}_t$ to the sparse output states $\vec{r} \in R_{\text{AN}}$. All similarities have been labelled with the state they represent and the colours are purely illustrative. The attractor transitions shown here are explicitly state-dependent, as can be seen from the repeated input of the stimulus ``father\_is'', which results in a transition to state ``Kronos'' when in ``Hades'', but to ``Uranus'' when in ``Kronos''. Additionally, the network is unaffected by nonsense input that does not correspond to a stored edge, as the network remains in the attractor ``Uranus'' when presented with the stimulus ``father\_is''.
}
\label{fig:walk_standard}
\end{figure*}

To show the generality of FSM construction, we chose to implement a directed graph representing the relationships between gods in ancient Greek mythology, due to the graph's dense connectivity. The graph and thus FSM to be implemented is shown in \mbox{\mbox{Figure \ref{fig:starwars_net}}}. From the graph it is clear that a state machine representing the graph must explicitly be capable of state-dependent transitions, e.g. the input ``overthrown\_by'' must result in a transition to state ``Kronos'' when in state ``Uranus'', but to state ``Zeus'' when in state ``Kronos''. To construct $\mat{W}$, the necessary hypervectors are first generated. For every state $\chi \in X_{\text{FSM}}$ in the FSM (e.g. ``Zeus'', ``Kronos'') a random bipolar hypervector $\vec{x}$ is generated according to Equation \ref{eqn:probX_half}. For every unique stimulus $\varsigma \in S_{\text{FSM}}$ (e.g. ``overthrown\_by'', ``father\_is'') a pair of random bipolar stimulus hypervectors $\vec{s}_a$ and $\vec{s}_b$ is likewise generated. Similarly, sparse ternary output hypervectors $\vec{r}$ are also generated. The weights matrix $\mat{W}$ is then iteratively constructed as per Equations \ref{eqn:W_construction} and \ref{eqn:tran_term_final}, with a new hypervector $\vec{e}$ also being generated for every edge. The matrix generated from this procedure we denote $\mat{W}^{\text{ideal}}$. For all of the following results, the attractor network is first initialised to be in a certain node attractor state, in this case, ``Hades''. The network is then allowed to freely evolve for 10 time steps (chosen arbitrarily) as per \mbox{Equation \ref{eqn:update_hopfield}}, with every neuron being updated simultaneously on every time step. During this period, it is desired that the network state $\vec{z}_t$ remains in the attractor state in which it was initialised. An input stimulus $\vec{s}_a$ is then presented to the network for 10 time steps, during which time the network state is masked by the stimulus hypervector, and the network evolves synchronously according to \mbox{Equation \ref{eqn:update_hop_mask}}. If the stimulus corresponds to a valid edge in the FSM, the network state $\vec{z}_t$ should then be driven towards the correct edge state attractor $\vec{e}$. After these 10 time steps, the second stimulus hypervector $\vec{s}_b$ for a particular input is presented for 10 time steps. Again, the network evolves according to \mbox{Equation \ref{eqn:update_hop_mask}}, and the network should be driven towards the target attractor state $\vec{y}$, completing the transition. This process is repeated every 30 time steps, causing the network state $\vec{z}_t$ to travel between node attractor states $\vec{x} \in X_{\text{AN}}$, corresponding to a valid walk between states $\chi \in X_{\text{FSM}}$ in the represented FSM. To view the resulting network dynamics, the similarity between the network state $\vec{z}_t$ and the edge- and node attractor states is calculated as per \mbox{Equation \ref{eqn:d_simple}}, such that a similarity of 1 between $\vec{z}_t$ and some attractor state $\vec{x}^\nu$ implies $\vec{z}_t = \vec{x}^\nu$ and thus that the network is inhabiting that attractor. The similarity between the network state $\vec{z}_t$ and the outputs states $\vec{r} \in R_{\text{AN}}$ is also calculated, but due to the output hypervectors being sparse, the maximum value that the similarity can take is $d(\vec{z}_t, \vec{r}) = f_r$, which would be interpreted as that output symbol being present.

An attractor network performing a walk is shown in \mbox{\mbox{Figure \ref{fig:walk_standard}}}, with parameters $N = 10,000$, $N f_r = 200$, $N_{Z} = 8$, and $N_E = 16$. This corresponds to the network having a per-neuron noise (the finite size effect resulting from random hypervectors having a nonzero similarity to each-other) of $\sigma \approx 0.07$, calculated via \mbox{Equation \ref{eqn:sigma_SNR}}. The magnitude of the noise is thus small compared with the desired signal of magnitude 1 (\mbox{Equation \ref{eqn:normal_attractor}}), and so we are far away from reaching the memory capacity of the network. The network performs the walk as intended, transitioning between the correct node attractor states and corresponding edge states with their associated outputs. The specific sequence of inputs was chosen to show the generality of implementable state transitions. First, there is the explicit state dependence in the repeated input of ``father\_is, father\_is''. Second, it contains an input stimulus that does not correspond to a valid edge for the currently inhabited state ( ``Zeus overthrown\_by''), which should not cause a transition. Third, it contains bidirectional edges ( ``consort\_is''), whose repeated application causes the network to flip between two states (between ``Kronos'' and ``Rhea''). And fourthly self-connections, whose target states and source states are identical. Since the network traverses all these edges as expected, we do not expect the precise structure of an FSM's graph to limit whether or not it can be emulated by the attractor network.

\subsection{Network robustness}
One of the advantages of attractor neural networks that make them suitable as plausible biological models is their robustness to imperfect weights \cite{amit_modeling_1989}. That is, individual synapses may have very few bits of precision or become damaged, yet the relevant brain region must still be able to carry out its functional task. To this end, we subjected the network presented here to similar non-idealities, to check that the network retains the feature of global stability and robustness despite being implemented with low-precision and noisy weights. In the first of these tests, the ideal weights matrix $\mat{W}^{\text{ideal}}$ was binarised and then additive noise was applied, via
\begin{equation}
    W^{\text{noisy}}_{ij} = \mathrm{sgn}\big( W^{\text{ideal}}_{ij}) + \sigma_{\text{noise}} \cdot \chi_{ij}
    \label{eqn:W_noisy}
\end{equation}
\begin{figure*}
\centering
\includegraphics[width=0.98\linewidth]{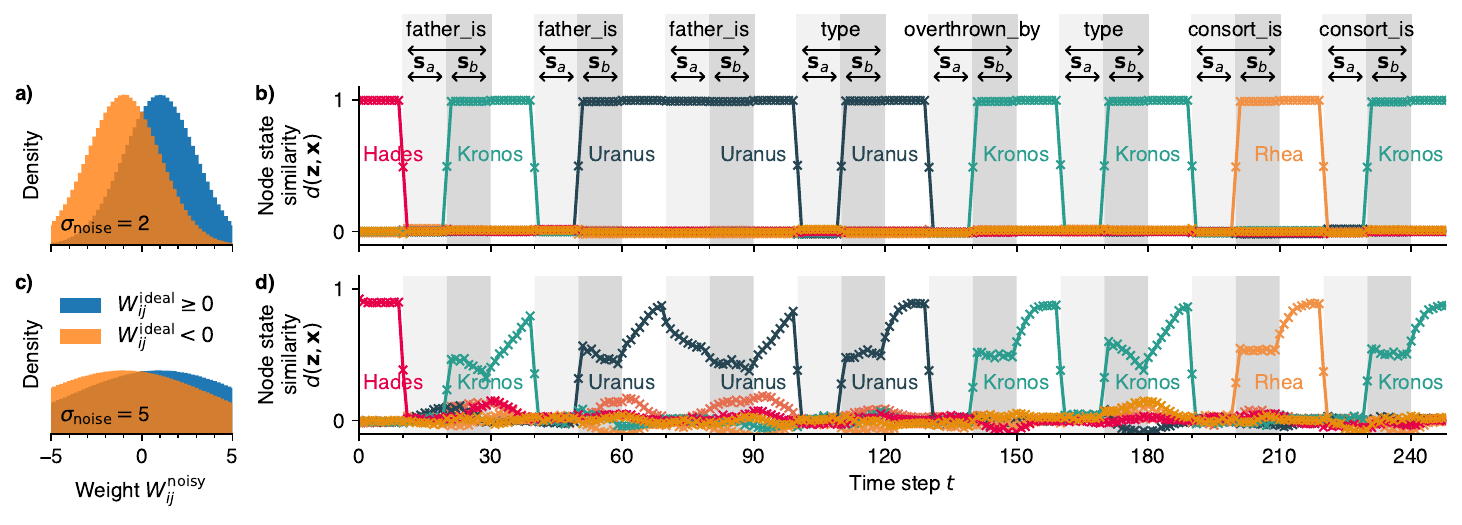}
\caption{The attractor network performing a walk as in \mbox{Figure \ref{fig:walk_standard}}, but using the damaged weights matrix $\mat{W}^{\text{noisy}}$, whose entries have been binarised and then independent additive noise has been applied, as per \mbox{Equation \ref{eqn:W_construction}}. \textbf{a)} The distribution of weights after they have been thusly damaged with noise of magnitude $\sigma_{\text{noise}} = 2$, corresponding to an SNR of \SI{0}{dB}. Weights whose ideal values were positive or negative have been plotted separately. \textbf{b)} The similarity of the network state $\vec{z}_t$ to stored node hypervectors, with the network using the weights from a). Shown above is the sequence of inputs given to the network, identical to in \mbox{Figure \ref{fig:walk_standard}}. \textbf{c)} The distribution of weights damaged with $\sigma_{\text{noise}} = 5$, corresponding to an SNR of \SI{-0.8}{dB}. \textbf{d)} The similarity of the network state to stored node hypervectors, but with the network using the damaged weights from c). The network transitions are thus highly robust to unreliable weights, and show a gradual degradation in performance, even when the network's weights are majorly imprecise and noisy. For both b) and d) the edge state and output similarity plots have been omitted for visual clarity.}
\label{fig:starwars_smear}
\end{figure*}
where $\chi_{ij} \in \mathbb{R}$ are independently sampled standard Gaussian variables, sampled once during matrix construction, and $\sigma_{\text{noise}} \in \mathbb{R}$ is a scaling factor on the strength of noise being imposed. The $\mathrm{sgn}(\cdot)$ function forces the weights to be bipolar, emulating that the synapses may have only 1 bit of precision, while the $\chi_{ij}$ random variables act as a smearing on the weight state, emulating that the two weight states have a finite width. A $\sigma_{\text{noise}}$ value of $2$ thus corresponds to the magnitude of the noise being equal to that of the signal (whether $W^{\text{ideal}}_{ij} \geq 0$), and so, for example, for a damaged weight value of $W^{\text{noisy}}_{ij} = +1$ there is a 38\% chance that the pre-damaged weight $W^{\text{ideal}}_{ij} = -1$. This level of degradation is far worse than is expected even from novel binary memory devices \cite{xia_memristive_2019}, and presumably also for biology. We used the same set of hypervectors and sequence of inputs as in \mbox{Figure \ref{fig:walk_standard}}, but this time using the degraded weights matrix $\mat{W}^{\text{noisy}}$, to test the network's robustness. The results are shown in \mbox{Figure \ref{fig:starwars_smear}} for weight degradation values of $\sigma_{\text{noise}} = 2$ and $\sigma_{\text{noise}} = 5$, corresponding to signal-to-noise ratios (SNRs) of \SI{0}{dB} and \SI{-0.8}{dB} respectively. We see that for $\sigma_{\text{noise}} = 2$ the attractor network performs the walk just as well as in Figure $\ref{fig:walk_standard}$, which used the ideal weights matrix, despite the fact that here the binary weight distributions overlap each-other considerably. Furthermore, we have that $d(\vec{z}_t, \vec{x}^\nu) \approx 1$ where $\vec{x}^\nu$ is the attractor that the network should be inhabiting at any time, indicating that the attractor stability and recall accuracy is unaffected by the non-idealities. For $\sigma_{\text{noise}} = 5$, a scenario where the realised weight carries very little information about the ideal weight's value, we see that the network nonetheless continues to function, performing the correct walk between attractor states. However, there is a degradation in the recall of stored attractor states, with the network state no longer converging to a similarity of 1 with the stored attractor states. For greater values of $\sigma_{\text{noise}}$, the network ceases to perform the correct walk, and indeed does not converge on any stored attractor state (not shown). 

A further test of robustness was to restrict the weights matrix to be sparse, as a dense all-to-all connectivity may not be feasible in biology, where synaptic connections are spatially constrained and have an associated chemical cost. Similar to the previous test, the sparse weights matrix was generated via
\begin{equation}
    W^{\text{sparse}}_{ij} = \mathrm{sgn}(W_{ij}^{\text{ideal}} \, ) \cdot \heavi (\, \lvert W_{ij} \rvert - \theta)
    \label{eqn:W_sparse}
\end{equation}
\begin{figure*}
\centering
\includegraphics[width=0.9\linewidth]{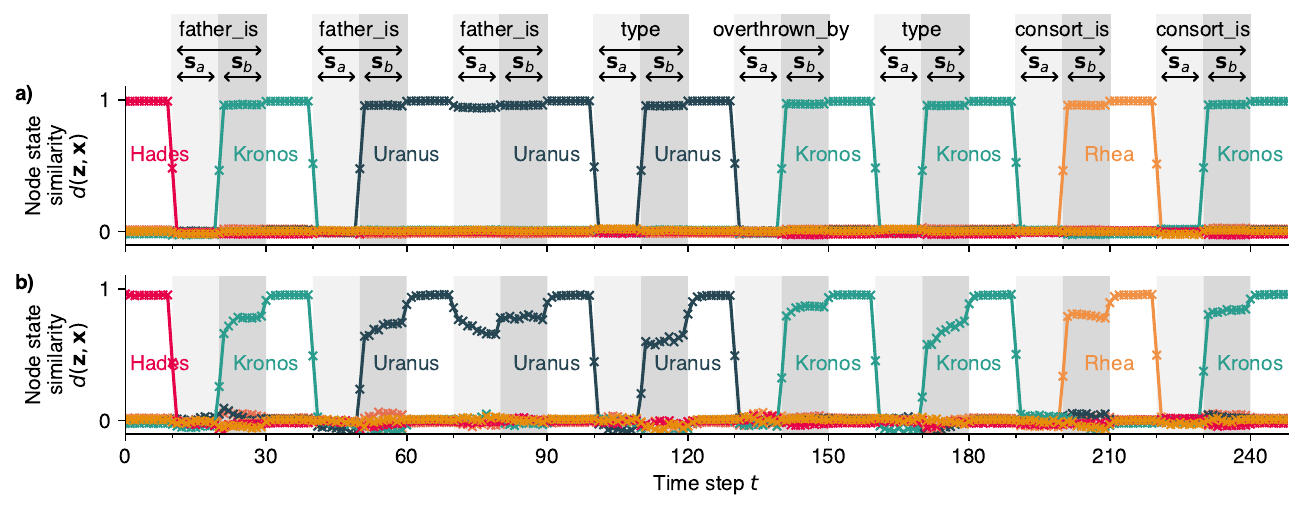}
\caption{The attractor network performing a walk as in \mbox{Figure \ref{fig:walk_standard}}, but using a sparse ternary weights matrix $\mat{W}^{\text{sparse}} \in \{-1,0,1\}^{N \times N}$, generated via \mbox{Equation \ref{eqn:W_sparse}}. The weights matrices for \textbf{a)} and \textbf{b)} are 98\% and 99\% sparse respectively. Shown are the similarities of the network state $\vec{z}_t$ with stored node hypervectors $\vec{x} \in X_{\text{AN}}$, with the applied stimulus hypervector at any time shown above. We see that even when 98\% of the entries in $\mat{W}$ are zeroes, the network continues to function with negligible loss in stability, as the correct walk between attractor states is performed, and the network converges on stored attractors with similarity $d(\vec{z}_t, \vec{x}) \approx 1$. At 99\% sparsity there is a degradation in the accuracy of stored attractors, with the network converging on states with $d(\vec{z}_t, \vec{x}) < 1$, but with the correct walk still being performed. Beyond 99\% sparsity the attractor dynamics break down (not shown). Thus although requiring a large number of neurons $N$ to enforce state pseudo-orthogonality, the network requires far fewer than $N^2$ nonzero weights to function robustly.}
\label{fig:starwars_sparse}
\end{figure*}
where $\theta$ is a threshold set such that $\mat{W}^{\text{sparse}} \in \{-1,0,1\}^{N \times N}$ has the desired sparsity. Through this procedure, only the most extreme weight values are allowed to be nonzero. Since the terms inside $\mat{W}^{\text{ideal}}$ are symmetrically distributed around 0, there are approximately as many +1 entries in $\mat{W}^{\text{sparse}}$ as -1s. Using the same hypervectors and sequence of inputs as before, an attractor network performing a walk using the sparse weights matrix $\mat{W}^{\text{sparse}}$ is shown in \mbox{Figure \ref{fig:starwars_sparse}}, with sparsities of 98\% and 99\%. We see that for the 98\% sparse case, there is again very little difference with the ideal case shown in \mbox{Figure \ref{fig:walk_standard}}, with the network still having a similarity of $d(\vec{z}_t, \vec{x}) \approx 1$ with stored attractor states, and performing the correct walk. When the sparsity is pushed further to 99\% however, we see that despite the network performing the correct walk, the attractor states are again slightly degraded, with the network converging on states with $d(\vec{z}_t, \vec{x}^\nu) < 1$ with stored attractor states $\vec{x}^\nu$. For greater sparsities, the network ceases to perform the correct walk, and again does not converge on any stored attractor state (not shown).

These two tests thus highlight the extreme robustness of the model to imprecise and unreliable weights. The network may be implemented with 1 bit precision weights, whose weight distributions are entirely overlapping, or set 98\% of the weights to 0, and still continue to function without any discernible loss in performance. The extent to which the weights matrix may be degraded and the network still remain stable is of course a function not only of the level of degradation, but also of the size of the network $N$, as well as the the number of FSM states $N_Z$ and edges $N_E$ stored within the network. For conventional Hopfield models with Hebbian learning, these two factors are normally theoretically treated alike, as contributing an effective noise to the postsynaptic sum as in \mbox{Equation \ref{eqn:sigma_SNR}}, and so the magnitude of withstandable synaptic noise increases with increasing $N$ \cite{amit_modeling_1989, sompolinsky_theory_1987}. Although a thorough mathematical investigation into the scaling of weight degradation limits is justified, as a first result we have here given numerical data showing stability even in the most extreme cases of non-ideal weights, and expect that any implementation of the network with novel devices would be far away from such extremities.

\subsection{Asynchronous updates}

\begin{figure*}
\centering
\includegraphics[width=0.9\linewidth]{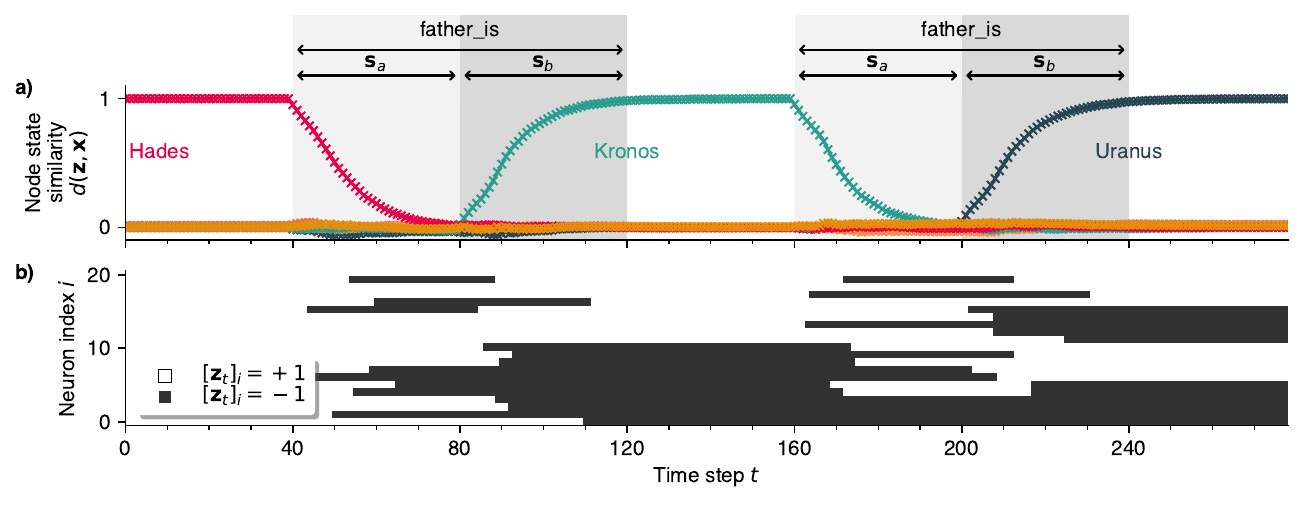}
\caption{An attractor network performing a shorter walk than in \mbox{Figure \ref{fig:walk_standard}}, but where neurons are updated asynchronously, with each neuron having a 10\% chance of updating on any time step. \textbf{a)} The similarity of the network state $\vec{z}_t$ to stored node hypervectors, with the stimulus hypervectors being applied to the network labelled above. \textbf{b)} The evolution of a subset of neurons within the attractor network, where for visual clarity, three of the node hypervectors have been taken from columns of the $N$-dimensional Hadamard matrix, rather than being randomly generated. The network functions largely the same as in the synchronous case, but with transitions between attractor states now taking a finite number of time steps to complete. The model is thus not dependent on the precise timing of neuron updates, and should function robustly in asynchronous systems where timing is unreliable.}
\label{fig:async}
\end{figure*}

Another useful property of Hopfield networks is the ability to robustly function even with asynchronously updating neurons, wherein not every neuron experiences a simultaneous state update. This property is especially important for any architecture claiming to be biologically plausible, as biological neurons update asynchronously and largely independent of each-other, without the the need for global clock signals. To this end, we ran a similar experiment to that in \mbox{Figure \ref{fig:walk_standard}}, using the undamaged weights matrix $\mat{W}^{\text{ideal}}$, but with an asynchronous neuron update rule, wherein on each time step every neuron has only a 10\% chance of updating its state. The remaining 90\% of the time, the neuron retains its state from the previous time step, regardless of its postsynaptic sum. There is thus no fixed order of neuron updates, and indeed it is not even a certainty that a neuron will update in any finite time. To account for the slower dynamics of the network state, the time for which inputs were presented to the network, as well as the periods without any input, was increased from 10 to 40 time steps. To be able to easily view the gradual state transition, three of the node hypervectors were chosen to be columns of the $N$-dimensional Hadamard matrix, rather than being randomly generated. The results are shown in \mbox{Figure \ref{fig:async}}, for a shorter sequence of stimulus inputs. We see that the network functions as intended, but with the network now converging on the correct attractors in a finite number of updates rather than in just one. The model proposed here is thus not reliant on synchronous dynamics, which is important not only for biological plausibility, but also when considering possible implementations on asynchronous neuromorphic hardware \cite{liu_event-based_2014, davies_loihi_2018}.

\subsection{Storage capacity}
It is well known that the storage capacity of a Hopfield network, the number of patterns $P$ that can be stored and reliably retrieved, is proportional to the size of the network, via $P < 0.14N$ \cite{hopfield_neural_1982, amit_modeling_1989}. When one tries to store more than $P$ attractors within the network, the so-called memory blackout occurs, after which no pattern can be retrieved.  We thus perform numerical simulations for a large range of attractor network and FSM sizes, to see if an analogous relationship exists. Said otherwise, for an attractor network of finite size $N$, what sizes of FSM can the network successfully emulate?

For a given $N$, number of FSM states $N_Z$ and edges $N_E$, a random FSM was generated and an attractor network constructed to represent it as described in Section \ref{section:methods}. To ensure a reasonable FSM was generated, the FSM's graph was first generated to have all nodes connected in a sequential ring structure, i.e. every state $\chi^{\nu}\in X_{\text{FSM}}$ connects to $\chi^{\nu+1 \mod N_Z}$. The remaining edges between nodes were selected at random, until the desired number of edges $N_E$ was reached. For each edge an associated stimulus is then required. Although one option would be to allocate as few unique stimuli as possible, so that the state transitions are maximally state-dependent, this results in some advantageous cancellation effects between the $\mat{E}^\eta$ transition terms and the stored attractors $\vec{x}^\nu \vec{x}^{\nu \intercal}$. To instead probe a worst-case scenario, each edge was assigned a unique stimulus.

\begin{figure*}
\centering
\includegraphics[width=5in]{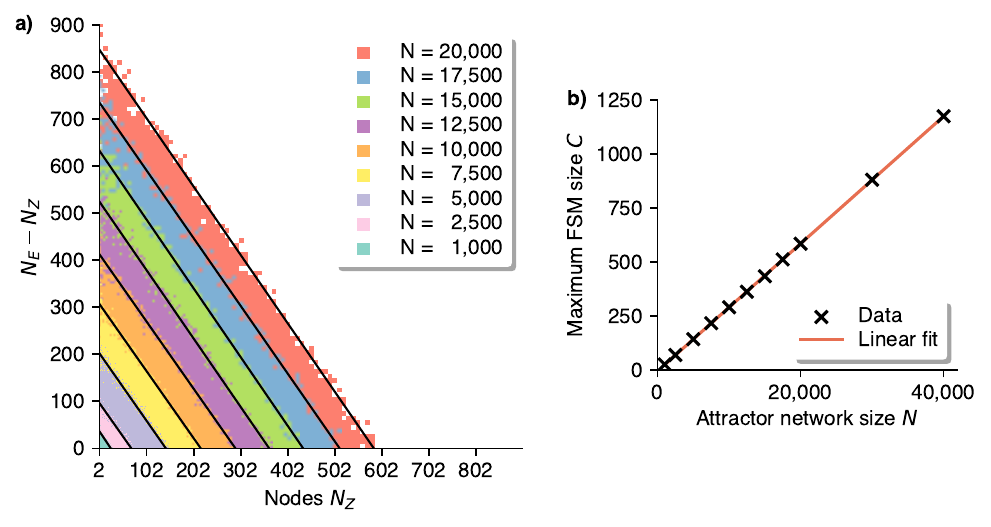}
\caption{The capacity of the attractor network for varying size $N$, in terms of the size of FSM that can be emulated before failure. For a given $N$, a random FSM was generated with number of nodes $N_Z$ and number of edges $N_E$. An attractor network was then constructed as described in Section \ref{section:methods}, and a sequence of stimuli input to the network that should trigger a specific walk between attractor states. \textbf{a)} Every coloured square is a successful walk, with no unique $(N_Z,N_E, N)$ triplet being sampled more than once, and lower-$N$ squares occlude higher-$N$ squares. Since only graphs with at least as many edges as nodes were sampled, $N_E - N_Z$ is given on the $y$-axis rather than $N_E$. The overlain black lines are the SVM-fitted decision boundaries, distinguishing between values that succeeded and values that failed. \textbf{b)} The capacity $C$ for varying Hopfield network sizes $N$, where $C$ is defined to be the maximum size of of FSM which can be implemented before failure, for which $N_E=N_Z$. A linear fit is overlain, and shows a linear relationship in the capacity $C$ in terms of $N$ over the range explored. Assuming that the gradients of the linear fit in a) are equal, the boundary at which failure and success are equiprobable is given by $N_Z + 2.2 N_E = 0.10 N$.}
\label{fig:memcap}
\end{figure*}

With the FSM now generated, an attractor network with $N$ neurons was constructed as previously described. An initial attractor state was chosen at random, and then a random valid walk between states was chosen to be performed (chosen arbitrarily to be of length 6, corresponding to each run taking 180 time steps). The corresponding sequence of stimuli was input to the attractor network via the same procedure as in \mbox{Figure \ref{fig:walk_standard}}, each masking the network state in turn. Each run was then evaluated to have either passed or failed, with a pass meaning that the network state inhabited the correct attractor state with overlap $d(\vec{z}_t, \vec{x}^\nu) > 0.5$ in the middle of all intervals when it should be in a certain node attractor state. This 0.5-criterion was chosen since, for a set of orthogonal hypervectors, at most only one hypervector may satisfy the criterion at once. A pass thus corresponds to the network performing the correct walk between attractor states. The results are shown in \mbox{Figure \ref{fig:memcap}}. We see that for a given $N$, there is a linear relationship between the the number of nodes $N_Z$ and number of edges $N_E$ in the FSM that can be implemented before failure. That this trade-off exists is not surprising, since both contribute additively to the SNR within the attractor network (\mbox{Equation \ref{eqn:sigma_SNR}}). For each $N$, a linear Support Vector Machine (SVM) was fitted to the data, to find the separating boundary at which failure and success of the walk are approximately equiprobable. The boundary is given by $N_Z + \beta N_E = c(N)$, where $\beta$ represents the relative cost of adding nodes and edges, and $c(N)$ is an offset. For all of the fitted boundaries, the value of $\beta$ was found to be approximately constant, with $\beta = 2.2 \pm 0.1$, and so is assumed to be independent of $N$. For every value of $N$, we define the capacity $C$ to be the maximum size of FSM which can be implemented before failure, for which $N_E = N_Z$. The capacity $C$ is then given by $C(N) = \frac{c(N)}{1+\beta}$, and is also plotted in \mbox{Figure \ref{fig:memcap}}.
A linear fit reveals an approximate proportionality relationship of $C(N) \approx 0.029 N$. Combining these two results, the boundary which limits the size of FSM which can be emulated is then given by 
\begin{equation}
    N_Z + 2.2 N_E < 0.10 N
\end{equation}
It is expected that additional edges consume more of the network's storage capacity than additional nodes, since for every edge, 5 additional terms are added to $\mat{W}$ (\mbox{Equation \ref{eqn:tran_term_final}}), contributing 3$\times$ as much cross-talk noise as adding a node would (\mbox{Equation \ref{eqn:sigma_SNR}}). We can compare this storage capacity relation with that of the standard Hopfield model, by considering the case $N_E = 0$, i.e. there are no transition terms in the network, and so the network is identical to a standard Hopfield network. In this case, our failure boundary would become $N_Z < 0.10N$, in comparison to Hopfield's $P < 0.14 N$.
% Although this may seem like a drastic reduction in memory capacity, we must remember that as a result of input stimuli applying a masking operation to our network, the effective size of the network during these periods is actually $N^\prime :=\frac{1}{2}N$. In this case, the failure boundary can be rewritten as $N_Z < 0.20 N^\prime$, which is more in keeping with the Hopfield estimate\footnote{Although it might seem we are claiming that the network implemented here has a greater storage capacity than standard Hopfield networks, our boundary at $0.20N$ is for equiprobable success and failure, while the $0.14N$ figure is given for overwhelmingly likely success.}.

\subsection{Storage capacity with sparse states}
\label{section:sparse_cap_results}

\begin{figure*}
\centering
\includegraphics[width=0.9\linewidth]{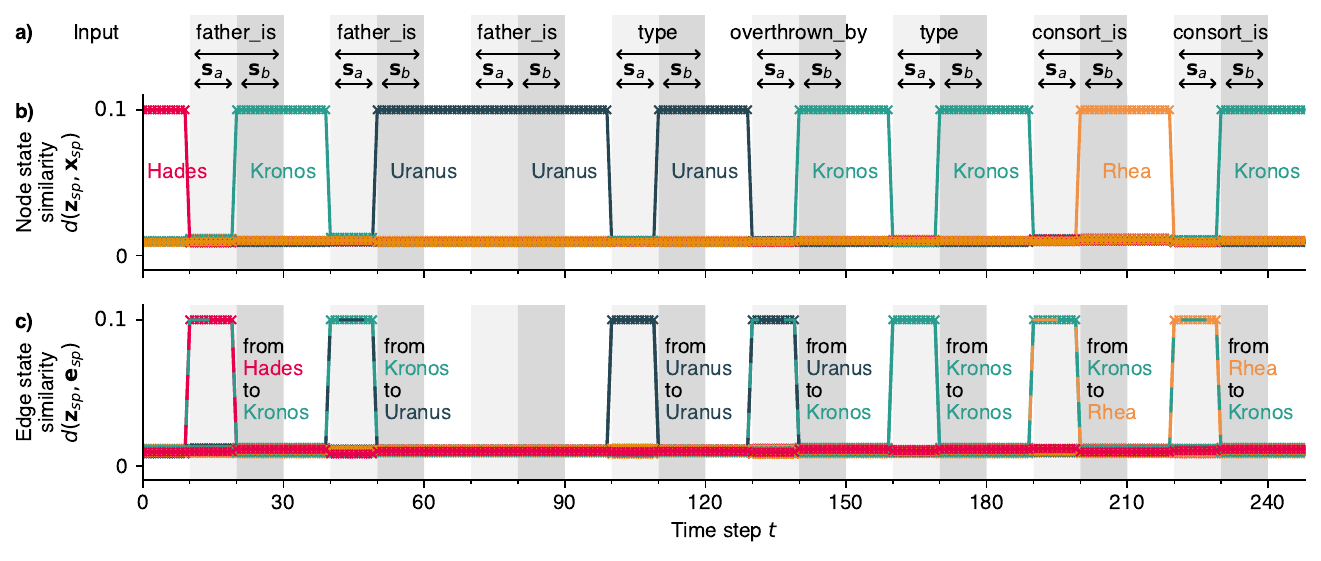}
\caption{The attractor network performing a walk between sparse attractor states, where the neurons have a top-$k$ binary activation function to enforce the desired sparsity (Equations \ref{eqn:activation_top_k} \& \ref{eqn:activation_top_k_mask}), and the weights matrix is constructed as discussed in Section \ref{section:sparse_theory}. The values used here are $N = 10,000$ neurons with coding level $f=0.1$, such that in any sparse hypervector $1000$ components are $+1$ while the rest are $0$. \textbf{a)} The input stimuli to the network, consisting of dense bipolar hypervectors applied as multiplicative masks. \textbf{b)} \& \textbf{c)} The overlap between the network state $\vec{z}_{t, \mathrm{sp}}$ to stored node attractor states $\vec{x}_\mathrm{sp}$ and stored edge attractor states $\vec{e}_\mathrm{sp}$ respectively, computed via the inner product (\mbox{Equation \ref{eqn:d_simple}}). Note that since the network and attractor states are now sparse binary, the maximum possible overlap value is $f=0.1$, while independently generated states have an expected overlap of $f^2 = 0.01$}
\label{fig:sparse_activity}
\end{figure*}

The same FSM as shown in \mbox{Figure \ref{fig:starwars_net}} was embedded into an attractor network via the construction scheme described in Section \ref{section:sparse_theory}, with values $N = 10,000$ neurons and coding level $f = 0.1$. To enforce the correct sparsity in the neural state, the $\mathrm{sgn}(\cdot )$ activation function \mbox{(Equation \ref{eqn:update_hopfield})} was replaced with a top-$k$ activation function (also known as ``$k$-Winners-Take-All'')
\begin{equation}
\label{eqn:activation_top_k}
\vec{z}_{t+1,  \mathrm{sp}} = H  \big( \mat{W} \vec{z}_{t, \mathrm{sp}} - \theta \big)
\end{equation}
where $H(\cdot)$ is a component-wise Heaviside function, and $\theta$ is chosen to be the $Nf$'th largest value of $\mat{W} \vec{z}_{t, \mathrm{sp}}$, to enforce that $\vec{z}_{t+1, \mathrm{sp}}$ is $f$-sparse. While a stimulus hypervector $\vec{s} \in \{-1,1\}^N$ is being applied as a mask to the network, the activation function is similarly
\begin{equation}
\label{eqn:activation_top_k_mask}
\vec{z}_{t+1, \mathrm{sp}} = H  \big(  \mat{W} (\vec{z}_{t, \mathrm{sp}}  \had H(\vec{s}))- \theta \big)
\end{equation}
with $\theta$ being chosen in the same manner. Note that although the introduction of this adaptive $\theta$ threshold mechanism may seem to be somewhat biologically implausible, or at least a tall order for any possible neural implementation, it may easily be implemented using a suitably connected population of inhibitory feedback neurons, which silence all attractor neurons except those that receive the greatest input \cite{amari_characteristics_1989, lin_sparse_2014}.
The sparse attractor network is shown performing a walk between the correct attractor states in \mbox{Figure \ref{fig:sparse_activity}}, as a sequence of stimuli is applied as input to the network. In contrast to the dense bipolar case, the maximum overlap between the network state $\vec{z}_{t, \mathrm{sp}}$ and a stored attractor state $\vec{x}^\nu_\mathrm{sp}$ is now $d(\vec{z}_{t, \mathrm{sp}}, \vec{x}^\nu_\mathrm{sp}) = f = 0.1$, while the expected overlap between unrelated states is $f^2 = 0.01$ rather than $0$.

We now apply the same procedure as in the dense case for determining the memory capacity of the sparse-activity attractor network. For direct comparison with the dense case, we define the memory capacity $C(N)$ to be the largest FSM with $N_E = N_Z$ for which walk success and failure are equiprobable. For every tested $(N, f, N_Z)$ tuple we generate a corresponding set of hypervectors and weights matrix as discussed in Section \ref{section:sparse_theory}, and then randomly choose a walk between 6 node attractor states to be completed. The chosen walk then determines the sequence of stimuli to be input, and each stimulus is then applied for $10$ time steps. Each $(N, f, N_Z)$ tuple was then determined to have passed or failed, with a success criterion that $d(\vec{z}_{t, \mathrm{sp}}, \vec{x}^\nu_\mathrm{sp}) > \frac{1}{2} (f + f^2)$ in the middle of all intervals when the network should be in a certain node attractor state. This criterion was chosen as it is the sparse analogue of that used in the dense case: at most only one attractor state may satisfy it at any time.

\begin{figure*}
\centering
\includegraphics[width=6in]{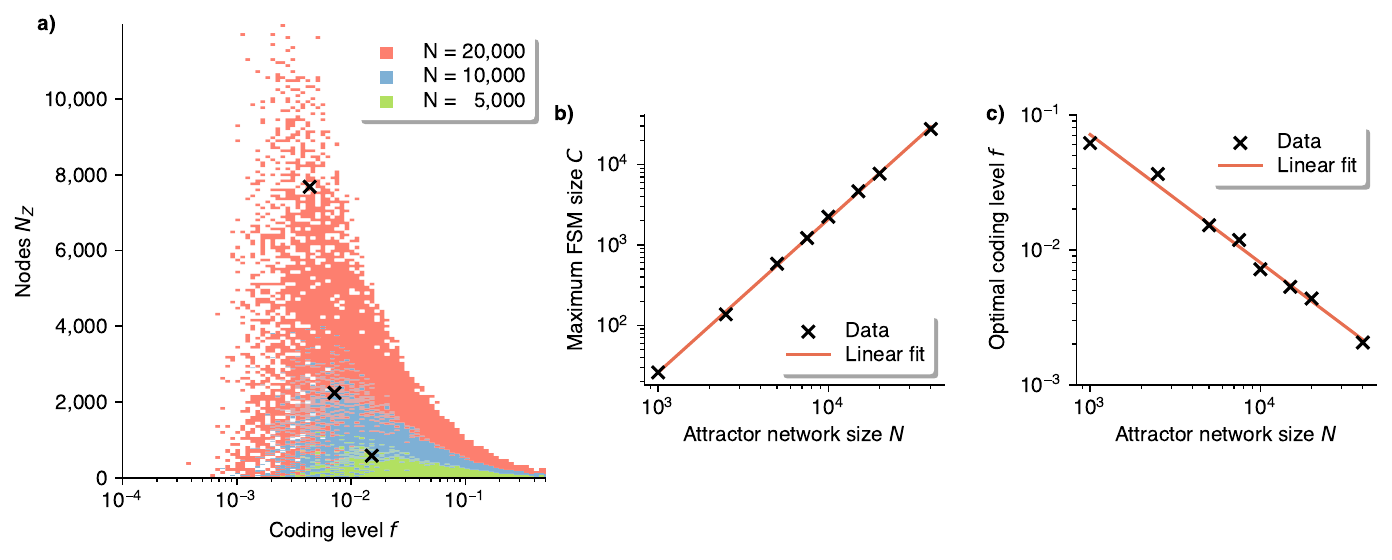}
\caption{The capacity of the attractor network with sparse binary activity and attractor states, for varying coding level $f$. \textbf{a)} Each coloured square is a successful walk, with no unique $(N, f, N_Z)$ tuple being tested more than once, and lower-$N$ squares occlude higher-$N$ squares for visual clarity. To comply with the definition of the memory capacity $C$ in the dense case, each FSM was generated with an equal number of states as edges, $N_Z = N_E$. The capacity $C$ is taken as the maximum $N_Z$ value for an $N$ at which the walk success probability $p_\mathrm{KDE} \geq 50\%$, estimated via a Gaussian KDE and indicated by the black crosses. \textbf{b)} The capacities $C$ obtained by this procedure for varying attractor network sizes $N$, up to $N = 40,000$, and \textbf{c)} the coding levels $f$ at these points. Linear fits are overlain for each, implying an approximately quadratic scaling relation for the memory capacity $C \sim N^{1.90}$ and an approximately inverse scaling relation for the coding level $f \sim N^{-0.949}$.}
\label{fig:memcap_sparse}
\end{figure*}

The results are shown in \mbox{Figure \ref{fig:memcap_sparse}}. We see that for a fixed number of neurons $N$, the size of FSM that may be stored initially increases as $f$ is decreased, but below a certain $f$ value drops off rapidly. To estimate the optimal coding level $f$ and maximum FSM size $N_Z$ for an attractor network of size $N$, we apply a 2D Gaussian convolutional filter with standard deviation 3 over the grid of successes/failures for each $N$ value separately, in order to obtain a kernel density estimate (KDE) $p_\mathrm{KDE}$ of the walk success probability. The capacity $C(N)$ was then obtained by taking the maximum $N_Z$ value for which $p_\mathrm{KDE} \geq 0.5$. This procedure was chosen in order to be comparable to that performed in the dense bipolar case \mbox{(Figure \ref{fig:memcap})}, where a linear separation boundary between success and failure was used instead. Plotting capacity $C$ against $N$ and applying a linear fit in the log-log domain reveals a scaling relation of $C \sim N^{1.90}$. This approximately quadratic scaling in the sparse case is a vast improvement over the linear scaling shown in the dense case \mbox{(Figure \ref{fig:memcap})}, and is in keeping with the theoretical scaling estimates of $P_{\mathrm{max}} \sim N^2 / (\log{N})^2$ for sparsely-coded binary attractor networks  \cite{amari_characteristics_1989}. The optimal coding level $f$ is also shown, and a linear fit in the log-log domain implies a scaling relation of the form $f \sim N^{-0.949}$. Again, this is similar to the theoretically optimal $f(N)$ scaling relation for sparse binary attractor networks, where the coding level scales like  $f \sim (\log{N}) / N$ \cite{amari_characteristics_1989}.

\section{Relation to other architectures}
\label{section:other_architectures}

\subsection{FSM emulation}

While there is a large body of work concerning the equivalence between RNNs and FSMs, their implementations broadly fall into a few categories. There are those that require iterative gradient descent methods to mimic an FSM \cite{zeng_learning_1993, lee_giles_learning_1995, das_unified_1994, pollack_induction_1991}, which makes them difficult to train for large FSMs, and improbable for use in biology. There are those that require creating a new FSM with an explicitly expanded state set, $Z' := Z \times S$, such that there is a new state for every old state-stimulus pair \cite{minsky_computation_1967, alquezar_algebraic_1995}, which is unfavourable due to the the explosion of (usually one-hot) states needing to be represented, as well as the difficulty of adding new states or stimuli iteratively. There are those that require higher-order weight tensors in order to explicitly provide a weight entry for every unique state-stimulus pair \cite{omlin_fuzzy_1998, forcada_finite-state_2001, mali_neural_2020} which, as well as being non-distributed, may be more difficult to implement, for example requiring the use of Sigma-Pi units \cite{koch_biophysics_1998, groschner_biophysical_2022} or a large number of hidden neurons with 2-body synaptic interactions only \cite{krotov_large_2021}.

In \oldcite{recanatesi_memory_2017} transitions are triggered by adiabatically modulating a global inhibition parameter, such that the network may transition between similar stored patterns. Lacking however is a method to construct a network to perform arbitrary, controllable transitions between states. In \oldcite{chen_attractor-state_2020} an in-depth analysis of small populations of rate-based neurons is conducted, wherein synapses with short-term synaptic depression enable a rich behaviour of itinerancy between attractor states, but does not scale to large systems and arbitrary stored memories.

Most closely resembling our approach, however, are earlier works concerned with the related task of creating a sequence of transitions between attractor states in Hopfield-like neural networks. The majority of these efforts rely upon the use of synaptic delays, such that the postsynaptic sum on a time step $t$ depends, for example, also on the network state at time $t-10$, rather than just $t-1$. These delay synapses thus allow attractor cross-terms of the form $\vec{x}^{\nu+1}\vec{x}^{\nu \intercal}$ to become influential only after the network has inhabited an attractor state for a certain amount of time, triggering a walk between attractor states \cite{sompolinsky_temporal_1986, kleinfeld_sequential_1986}. This then also allowed for the construction of networks with state-dependent input-triggered transitions \cite{gutfreund_processing_1988, amit_neural_1988, drossaers_hopfield_1992}. Similar networks were shown to function without the need for synaptic delays, but require fine tuning of network parameters and suffer from extremely low storage capacity \cite{buhmann_noise-driven_1987, amit_modeling_1989}. In any case, the need for synaptic delay elements represents a large requirement on any substrate which might implement such a network, and indeed are problematic to implement in neuromorphic systems \cite{nielsen_compact_2017}.

State-dependent computation in spiking neural networks was realised in \oldcite{neftci_synthesizing_2013} and \oldcite{liang_neural_2019}, where they used population attractor dynamics to achieve robust state representations via sustained spiking activity. Additionally, these works highlight the need for robust-yet-flexible neural state machine primitives, if one is to succeed in designing intelligent end-to-end neuromorphic cognitive systems. These approaches differ from this work however in that the state representations are still fundamentally population-based rather than distributed, and so pose difficulties such as the requirement of finding a new population of neurons to represent any new state \cite{rutishauser_state-dependent_2009}.

In \oldcite{rigotti_internal_2010} they discuss the need for a mechanism to induce flips in the neuron state (i.e. an operation akin to a Hadamard product) in order to directly implement nontrivial switching between different attractor states, but disqualify such a mechanism from plausibly existing using synaptic currents alone. We also reject such a mechanism as a biologically plausible solution, but on the grounds that it would not robustly function in an asynchronous neural system (see Section \ref{subsec:hadamard_input}).
They instead show the necessity of a population of neurons with mixed selectivity, connected to both the input and attractor neurons, in order to achieve the desired attractor itinerancy dynamics.
% The lack of a flipping mechanism is also discussed in \oldcite{rigotti_internal_2010}, wherein they show the necessity of a population of neurons with mixed selectivity, connected to both the input and attractor neurons, in order to achieve the flipping-like behaviour necessary for complex state switching.
This requirement arose by demanding that the network state switch to resembling the target state immediately upon receiving a stimulus. We instead show that similar results can be achieved without this extra population, if we relax to instead demanding only that the network soon evolve to the target state.

The main contribution of this article is thus to introduce a method by which attractor networks may be endowed with state-dependent attractor-switching capabilities, without requiring biologically implausible elements or components which are expensive to implement (e.g. precise synaptic delays), and can be scaled up efficiently. The extension to arbitrary FSM emulation shows the generality of the method, and that its limitations can be overcome by the appropriate modifications, like introducing the edge state attractors (Section \ref{sec:why_edge_states}).

\subsection{VSA embeddings}

This work also differs from more conventional methods to implement graphs and FSMs in VSAs \cite{poduval_graphd_2022, kleyko_survey_2022, yerxa_hyperdimensional_2018, osipov_associative_2017, teeters_separating_2023}, in that the network state does not need to be read by an outsider in order to implement the state transition dynamics. That is, where in previous works a graph is encoded by a hypervector (or an associative memory composed of hypervectors) such that the desired dynamics and outputs may be reliably decoded by external circuitry, we instead encode the graph's connectivity within the attractor network's weights matrix, such that its recurrent neural dynamics realise the desired state machine behaviour. 

The use of a Hopfield network as an auto-associative cleanup memory in conjunction with VSAs has been explored in previous works, including theoretical analyses of their capacity to store bundled hypervectors with different representations \cite{clarkson_capacity_2023}, and using single attractor states to retrieve knowledge structures from partial cues \cite{steinberg_associative_2022}. Further links between VSAs and attractor networks have also been demonstrated with the use of complex phasor hypervectors - rather than binary or bipolar hypervectors - being stored as attractors within phasor neural networks \cite{plate_holographic_2003, kleyko_survey_2022, frady_robust_2019, noest_phasor_1987}. Complex phasor hypervectors are of particular interest in neuromorphic computing, since they may be very naturally implemented with spike-timing phasor codes, wherein the value represented by a neuron is encoded by the precise timing of its spikes with respect to other neurons or a global oscillatory reference signal, and hypervector binding may be implemented by phase addition \cite{auge_survey_2021, orchard_hyperdimensional_2023}.

In \oldcite{osipov_associative_2017} the authors show the usefulness of VSA representations for synthesizing state machines from observable data, which might be combined with this work to realise a neural system that can synthesise appropriate attractor itinerancy dynamics to best fit observed data.
Similarly, if equally robust attractor-based neural implementations of other primitive computational blocks could be created - such as a stack - then they might be combined to create more complex VSA-driven cognitive computational structures, such as neural Turing machines \cite{yerxa_hyperdimensional_2018, graves_neural_2014, grefenstette_learning_2015}. Looking further, this combined with the end-to-end trainability of VSA models could pave the way for neural systems which have the explainability, compositionality and robustness thereof, but the flexibility and performance of deep neural networks \cite{hersche_neuro-vector-symbolic_2023, schlag_enhancing_2020}.

\section{Biological plausibility}

Transitions between discrete neural attractor states are thought to be a crucial mechanism for performing context-dependent decision making in biological neural systems \cite{daelli_neural_2010, mante_context-dependent_2013, miller_itinerancy_2016, tajima_task-dependent_2017}. Attractor dynamics enable a temporary retention of received information, and ensure that irrelevant inputs do not produce stable deviations in the neural state.  Such networks are widely theorised to exist in the brain, for example in the hippocampus for its pattern completion and working memory capabilities \cite{rolls_mechanisms_2013, khona_attractor_2022}. As such, we showed that a Hopfield attractor network and its sparse variant can be modified such that they can perform stimulus-triggered state-dependent attractor transitions, without resorting to additional biologically-implausible mechanisms and while abiding by the principles of distributed representation.
The changes we introduced are a) an altered weights matrix construction with additional asymmetric cross-terms (which does not incur any considerable extra complexity) and b) the ability for a stimulus to mask a subset of neurons within the attractor population. As long as such a mechanism exists, the network proposed here could thus map onto brain areas theorised to support attractor dynamics. The masking mechanism could, for example, feasibly be achieved by a population of inhibitory neurons representing the stimuli, which selectively project to neurons within the attractor population.

\subsection{Robustness}

The robust functioning of the network despite noisy and unreliable weights is a crucial prerequisite for the model to plausibly be able to exist in biological systems. As we have shown, the network weights may be considerably degraded without affecting the behaviour of the network, and indeed beyond this the network exhibits a so-called graceful degradation in performance. Furthermore, biological synapses are expected to have only a few bits of precision \cite{oconnor_graded_2005, bartol_nanoconnectomic_2015, baldassi_learning_2016}, and the network has been shown to function even in the worst case of binary weights. These properties stem from the massive redundancy arising from storing the attractor states across the entire synaptic matrix in a distributed manner, a technique that the brain is expected to utilise \cite{rumelhart_parallel_1986, crawford_biologically_2016}. Of course, we expect there to be a trade-off between the amount of each non-ideality that the network can withstand before failure. That is, an attractor network with dense noisy weights may withstand a greater degree of synaptic noise than if the weights matrix were also made sparse. Likewise, larger networks storing the same sized FSM should be able to withstand greater non-idealities than smaller networks, as is the case for attractor networks in general \cite{sompolinsky_theory_1987, amit_modeling_1989}.

Since the network is still an attractor network, it retains all of the properties that make them suitable for modelling cognitive function, such as that the network can perform robust pattern completion and correction, i.e. the recovery of a stored prototypical memory given a damaged, incomplete or noisy version, and thereafter function as a stable working memory \cite{hopfield_neural_1982, amit_modeling_1989}.

The robustness of the network to weight non-idealities also makes it a prime candidate for implementation with novel memristive crossbar technologies, which would allow an efficient and high-density implementation of the matrix-vector multiplication required in the neural state update \mbox{(Equation \ref{eqn:update_hop_mask})} to be performed in one operation  \cite{verleysen_analog_1989, ielmini_-memory_2018, xia_memristive_2019}. Akin to the biological synapses they emulate, such devices also often have only a few bits of precision, and suffer from considerable per-device mismatch in the programmed conductance states. The network proposed in this article is thus highly suitable for implementation with such architectures, as we have shown that robust performance is retained even when the network is subjected to very high degree of such non-idealities.

The continued functionality of the network when its dynamics are asynchronous is another important factor when considering its biological plausibility. In a biological neural system, neurons will produce action potentials whenever their membrane potential happens to exceed the neuron's spiking threshold, rather than all updating synchronously at fixed time intervals. We tested the regime where the timescale of the neuron dynamics is much slower than the timescale of the input, by replacing the synchronous neuron update rule with a stochastic asynchronous variant thereof, and showed that the network is robust to this asynchrony. Similarly, we tested the regime where neuron dynamics are much faster than the input, by considering input which is applied stochastically and asynchronously instead (Section \ref{subsec:hadamard_input}). The continued robustness of the model in these two extreme asynchronous regimes implies that the network is dependent neither upon the exact timing of inputs to the network, nor on the neuron updates within the network, and so would function robustly both in biological neural systems and asynchronous neuromorphic systems where the exact timing of events cannot be guaranteed \cite{liu_event-based_2014, davies_loihi_2018}.

\subsection{Learning}

The procedure for generating the weights matrix $\mat{W}$, as a result of its simplicity, makes the proposed network more biologically plausible than other more complex approaches, e.g. those utilising gradient descent methods. It can be learned in one-shot in a fully online fashion, since adding a new node or edge involves only an additive contribution to the weights matrix, which does not require knowledge of irrelevant edges, nodes, their hypervectors, or the weight values themselves. Furthermore, as a result of the entirely distributed representation of states and transitions, new behaviours may be added to the weights matrix at a later date, both without having to allocate new hardware, and without having to recalculate $\mat{W}$ with all previous data. Both of these factors are critical for continual online learning.

Evaluating the local learnability of $\mat{W}$ to implement transitions is also necessary to evaluate the biological plausibility of the model. In the original paper by Hopfield, the weights could be learned using the simple Hebbian rule
\begin{equation}
    \delta w_{ij} = x^{\nu}_{i} x^{\nu}_{j}
\end{equation}
where $x_i^\nu$ and $x_j^\nu$ are the activities of the post- and presynaptic neurons respectively, and $\delta w_{ij}$ the online synaptic efficacy update \cite{hebb_organization_1949, hopfield_neural_1982}. While the attractor terms within the network can be learned in this manner, the transition cross-terms that we have introduced require an altered version of the learning rule. If we simplify our network construction by removing the edge state attractors, then the local weight update required to learn a transition between states is given by
\begin{equation}
\label{eqn:dw_no_edge}
    \delta w_{ij} = H(s_i) y_{i} x_{j} s_{j}
\end{equation}
where $\vec{y}$, $\vec{x}$ and $\vec{s}$ are as previously defined. In removing the edge states, we disallow FSMs with consecutive edges with the same stimulus (e.g. ``father\_is, father\_is''), but this is not a problem if completely general FSM construction is not the goal per se (see Section \ref{sec:why_edge_states}, \mbox{Figure \ref{fig:pointy_carrot})}. This state-transition learning rule is just as local as the original Hopfield learning rule, as the weight update from presynaptic neuron $j$ to postsynaptic neuron $i$ is dependent only upon information that may be made directly accessible in the pre- and postsynaptic neurons, and does not depend on information in other neurons to which the synapse is not connected \cite{zenke_brain-inspired_2021, khacef_spike-based_2022}.

From the hardware perspective, the locality of the learning rule means that if the matrix-vector multiplication step in the neuron state update rule is implemented using novel memristive crossbar circuits  \cite{ielmini_-memory_2018, xia_memristive_2019, zidan_chapter_2020}, then the weights matrix could be learned online and in-memory via a sequence of parallel conductance updates, rather than by computing the weights matrix offline and then writing the summed values to the devices' conductances. As long as the updates in the memristors' conductances are sufficiently linear and symmetric, then attractors and transitions could be sequentially learned in one-shot and in parallel by specifying the two hypervectors in the outer product weight update at the crossbar's inputs and outputs by appropriately shaped voltage pulses \cite{alibart_pattern_2013, li_situ_2021}.

\subsection{Scaling}
\label{section:scaling}

When the FSM states are represented by dense bipolar hypervectors within the attractor network, we found a linear scaling between the size of the network $N$ and the capacity $C$ in terms of the size of FSM that could be embedded without errors. Although this is in keeping with the results in the Hopfield paper, this is not a favourable result when considering the biological plausibility of the system for large $N$ \cite{hopfield_neural_1982}. Since the attractor network is fully connected, the capacity actually scales sublinearly $C \sim \sqrt{N_\mathrm{syn}}$ with the number of synapses $N_\mathrm{syn}$, meaning that an increasing number of synapses are required per attractor and transition to be stored for large $N$, and so the network becomes increasingly inefficient. Additionally, the fact that every neuron is active at any time (or half of them, depending on interpretation of the $-1$ state) represents an unnecessarily large energy burden for any system utilising this model. This is in contrast to data from neural recordings, where a low per-neuron mean activity is ensured by the sparse coding of information \cite{rolls_neuronal_2011, olshausen_sparse_2004, barth_experimental_2012}.

We thus tested how the capacity of the network scales with $N$ when the FSM states are instead represented by sparse binary hypervectors with coding level $f$, since it is well known that the number of sparse binary vectors that can be stored in an attractor network scales much more favourably, $P \sim N^2 / (\log N)^2$ \cite{amari_characteristics_1989}. We found indeed that the sparse coding of the FSM states vastly improved the capacity of the network, scaling approximately quadratically with $C \sim N^{1.90}$, and so approximately linearly in the number of synapses. This linear scaling with the number of synapses ensures not only the efficient use of available synaptic resources in biological systems, but is especially important when one considers a possible implementation in neuromorphic hardware, where the number of synapses usually represents the main size constraint, rather than the number of neurons \cite{davies_loihi_2018, manohar_hardwaresoftware_2022}.

The coding level $f$ was found to have an approximately inverse relationship with the attractor network size, $f \sim N^{-0.949}$, which would imply that the number of active neurons $Nf$ in any attractor state grows very slowly, $Nf \sim N^{0.051}$. This is in agreement with the theoretically optimal case, where the coding level for a sparse binary attractor network should scale like $f \sim (\log N)/N$, and so the number of active neurons in any pattern scales like $Nf \sim \log N$ \cite{amari_characteristics_1989}.

Sparsity in the stored hypervectors is especially important when one considers how the weights matrix $\mat{W}$ could be learned in an online fashion, if the synapses are restricted to have only a few bits of precision. So far we have considered quantisation of the weights only \textit{after} the summed values have been determined, whereas including weight quantisation while new patterns are being iteratively learned is a much harder problem, and implies attractor capacity relations as poor as $P \sim \log N$. One solution is for the states to be increasingly sparse, in which case the optimal scaling of $P \sim N^2 / (\log N)^2$ can be recovered \cite{amit_learning_1994, brunel_slow_1998}.

In short, by letting the FSM states be represented by sparse binary hypervectors rather than dense bipolar hypervectors, we not only move closer to a more biologically realistic model of neural activity, but also benefit from the superior scaling properties of sparse binary attractor networks, which lets the maximum size of FSM that can be embedded scale approximately quadratically with the attractor network size rather than linearly.

\section{Conclusion}
\label{section:conclusion}

Attractor neural networks are robust abstract models of human memory, but previous attempts to endow them with complex and controllable attractor-switching capabilities have suffered mostly from being either non-distributed, not scalable, or not robust. We have here introduced a simple procedure by which any arbitrary FSM may be embedded into a large-enough Hopfield-like attractor network, where states and stimuli are represented by high-dimensional random hypervectors, and all information pertaining to FSM transitions is stored in the network's weights matrix in a fully distributed manner. Our method of modelling input to the network as a masking of the network state allows cross-terms between attractors to be stored in the weights matrix in a way that they are effectively obfuscated until the correct state-stimulus pair is present, much in a manner similar to the standard binding-unbinding operation in more conventional VSAs.

We showed that the network retains many of the features of attractor networks which make them suitable for biology, namely that the network is not reliant on synchronous dynamics and is robust to unreliable and imprecise weights, thus also making it highly suitable for implementation with high-density but noisy devices. We presented numerical results showing that the network capacity in terms of implementable FSM size scales linearly with the size of the attractor network for dense bipolar hypervectors, and approximately quadratically for sparse binary hypervectors.

In summary, we introduced an attractor-based neural state machine which overcomes many of the shortcomings that made previous models unsuitable for use in biology, and propose that attractor-based FSMs represent a plausible path by which FSMs may exist as a distributed computational primitive in biological neural networks.

% \section*{Acknowledgements}
% We thank Dr. Federico Corradi, Dr. Nicoletta Risi and Dr. Matthew Cook for their invaluable input and suggestions, as well as their help with proofreading this document. 

% Funded by the Deutsche Forschungsgemeinschaft (DFG, German Research Foundation) - Project MemTDE Project number 441959088 as part of the DFG priority program SPP 2262 MemrisTec - Project number 422738993, and Project NMVAC - Project number 432009531.
% %
% The authors would like to acknowledge the financial support of the CogniGron research center and the Ubbo Emmius Funds (Univ. of Groningen).

% \FloatBarrier

\printbibliography

\newpage

\onecolumn

\section*{Appendix}

\subsection{Dynamics without masking}
\label{section:appendix_no_mask}

For the following calculations we assume that the coding level of the output states $f_r$ is low enough that their effect can be ignored. With this in mind, if we ignore the semantic differences between attractors for node states and attractors for edge states, the two summations over states can be absorbed into one summation over both types of attractor, here both denoted $\vec{x}^\nu$. Similarly there is then no difference between the two transition cross-terms within each $\mat{E}$ term, and they too can be absorbed into one summation. Our simplified expression for $\mat{W}$ is now given by
\begin{equation}
    \mat{W} = \frac{1}{N}\sum_{\vphantom{\text{dg}}\text{attr's $\nu$}}^{N_Z + N_E} \vec{x}^\nu \vec{x}^{\nu\, \intercal}  +
    \frac{1}{N}\sum_{\text{tran's $\lambda$}}^{2N_{E}} \heavi(\vec{s}^{\pi(\lambda)}) \had (\vec{x}^{\upsilon(\lambda)} - \vec{x}^{\chi(\lambda)}) (\vec{x}^{\chi(\lambda)} \had \vec{s}^{\pi(\lambda)})\T
\label{eqn:apx_1}
\end{equation}
where $\chi(\lambda)$ and $\upsilon(\lambda)$ are functions $\{1,\ldots, 2N_E\} \rightarrow \{1,\ldots,N_Z + N_E \}$ determining the indices of the source and target states for transition $\lambda$, and $\pi(\lambda) : \{1, \ldots, 2N_E\} \rightarrow \{1, \ldots N_{\text{stimuli}}\}$ determines the index of the associated stimulus. We then wish to calculate the statistics of the postsynaptic sum $\mat{W}\vec{z}$ while the attractor network is currently in an attractor state. When in an attractor state $\vec{x}^\mu$, the postsynaptic sum is given by
%
% \begin{widetext}
\begin{equation}
\begin{split}
\Big[ \mat{W}\vec{x}^\mu \Big]_i & = \frac{1}{N}\sum_{\vphantom{\text{dg}}\text{attr's $\nu$}}^{N_Z + N_E} x_i^{\nu} \underbrace{\big[ \vec{x}^{\nu} \cdot \vec{x}^\mu \big]}_{ \substack{N\text{ if }  \mu = \nu \\ \text{  else  }   \mathcal{N}(0,N)}} 
      + \frac{1}{N}\sum_{\text{tran's $\lambda$}}^{2N_{E}} \heavi(s_i^{\pi(\lambda)}) \had (x_i^{ \upsilon (\lambda)} - x_i^{ \chi (\lambda)}) \underbrace{\big[ (\vec{x}^{ \chi (\lambda)} \had \vec{s}^{ \pi(\lambda)}) \cdot \vec{x}^\mu \big]}_{  \mathcal{N}(0,N)} \\
     & = x^\mu_i + \sum_{\vphantom{\text{dg}}\substack{\text{attr's}\\ \nu \neq \mu}}^{N_Z + N_E} \underbrace{x_i^{\nu}}_{ \mathrm{Var.} = 1 } \Big[  \, \mathcal{N}^{\nu} \big( 0,\frac{1}{N} \big) \Big] 
      + \sum_{\text{tran's $\lambda$}}^{2N_{E}} \underbrace{\heavi(s_i^{ \pi(\lambda)}) \had (x_i^{ \upsilon (\lambda)} - x_i^{ \chi (\lambda)})}_{ \mathrm{Var.}=1  }\Big[ \mathcal{N}^{\lambda} \big(0,\frac{1}{N} \big) \Big] \\
    & \approx x^\mu_i + \mathcal{N}\Bigg( 0, \frac{N_Z + N_E -1}{N} \Bigg) + \mathcal{N}\Bigg( 0, \frac{2N_E}{N} \Bigg) \\
    & \approx x^\mu_i + \mathcal{N}\Bigg( 0, \frac{N_Z + 3N_E}{N} \Bigg)
\end{split}
\label{eqn:apx_2}
\end{equation}
% \end{widetext}
%
where we have used the notation $\mathcal{N}(\mu, \sigma^2)$ to denote a normally-distributed random variable (RV) with mean $\mu$ and variance $\sigma^2$. In the third line we have made the approximation in the transition summation that the linear sum of attractor hypervectors, each multiplied by a Gaussian RV, is itself a separate Gaussian RV in each dimension. This holds as long as there are ``many'' attractor terms appearing on the LHS of the transition summation. Said otherwise, if the summation over transition terms has only very few unique attractor terms on the LHS ($N_E \gg N_Z$), then the noise will be a random linear sum of the same few (masked) hypervectors, each with approximate magnitude $\frac{1}{\sqrt{N}}$, and so will be highly correlated between dimensions. Nonetheless we assume we are far away from this regime, and let the effect of the sum of these unwanted terms be approximated by a normally-distributed random vector, and so we have
\begin{equation}
    \mat{W}\vec{x}^\mu \approx \vec{x}^\mu + \sigma\vec{n}
\end{equation}
where $\sigma =  \sqrt{\frac{N_Z + 3N_E}{N}}$ is the strength of cross-talk noise, and $\vec{n}$ a vector composed of IID standard normally-distributed RVs. This procedure of quantifying the signal-to-noise ratio (SNR) is adapted from that in the original Hopfield paper \cite{hopfield_neural_1982, amit_modeling_1989}.

\subsection{Dynamics with masking}
\label{section:appendix_mask}

We can similarly calculate the postsynaptic sum when in an attractor state $\vec{x}^\mu$, while the network is being masked by a stimulus $\vec{s}^\kappa$, with this (state, stimulus) tuple corresponding to a certain valid transition $\lambda^\prime$, with source, target, and stimulus hypervectors $\vec{x}^\mu$, $\vec{x}^\phi$, and $\vec{s}^\kappa$ respectively:
%
% \begin{strip}
\begin{equation}
\begin{split}
\Big[ \mat{W} \big( \vec{x}^\mu \had \heavi(\vec{s}^\kappa) \big)  \Big]_i & = \frac{1}{N}\sum_{\vphantom{\text{dg}}\text{attr's $\nu$}}^{N_Z + N_E} x_i^{\nu} \underbrace{\big[ \vec{x}^{\nu} \cdot \big( \vec{x}^\mu \had \heavi (\vec{s}^\kappa) \big) \big]}_{ \substack{\frac{1}{2}N\text{ if }  \mu = \nu \\ \text{  else  }   \mathcal{N}(0,\frac{1}{2}N)}} \\
    &  \quad \quad + \frac{1}{N}\sum_{\text{tran's $\lambda$}}^{2N_{E}} \heavi(s_i^{ \pi(\lambda)})  (x_i^{ \upsilon (\lambda)} - x_i^{ \chi (\lambda)}) \underbrace{\big[ (\vec{x}^{ \chi(\lambda)} \had \vec{s}^{ \pi(\lambda)}) \cdot \big( \vec{x}^\mu  \had \heavi (\vec{s}^\kappa) \big)\big]}_{ \substack{\frac{1}{2}N\text{ if }   \chi(\lambda) = \mu \text{ and } \pi(\lambda) = \kappa \\ \text{  else  }   \mathcal{N}(0,\frac{1}{2}N)}} \\
    & = \frac{1}{2}x^\mu_i + \frac{1}{2}\heavi(s_i^\kappa) (x_i^\phi - x_i^\mu) +  \sum_{\vphantom{\text{dg}}\substack{\text{attr's}\\ \nu \neq \mu}}^{N_Z + N_E} \underbrace{x_i^{\nu}}_{ \mathrm{Var.} = 1 } \Big[  \, \mathcal{N}^{\nu}(0,\frac{1}{2N}) \Big] \\
    & \quad \quad  + \sum_{\substack{\text{tran's} \\  \lambda \neq \lambda^\prime}}^{2N_{E}} \underbrace{\heavi(s_i^{ \pi(\lambda)}) (x_i^{ \upsilon(\lambda)} - x_i^{ \chi (\lambda)})}_{ \mathrm{Var.} = 1 } \Big( \mathcal{N}^{\lambda}(0,\frac{1}{2N}) \Big) \\
    & \approx \frac{1}{2} \big[ \heavi(s_i^\kappa) + \heavi(-s_i^\kappa) \big] x^\mu_i + \frac{1}{2}\heavi(s_i^\kappa)( x_i^\phi - x_i^\mu) \\
    & \quad \quad  + \mathcal{N}\Big( 0, \frac{N_Z + N_E -1}{2N} \Big) + \mathcal{N}\Big( 0, \frac{2N_E-1}{2N} \Big) \\
    % & = \frac{1}{2}\heavi(s_i^\kappa) x_i^\phi + \frac{1}{2}\heavi(-s_i^\kappa) x_i^\mu + \mathcal{N}\Big( 0, \frac{N_Z + N_E -1}{2N} \Big) + \mathcal{N}\Big( 0, \frac{2N_E-1}{2N} \Big) \\
    & = \frac{1}{2}\heavi(s_i^\kappa) x_i^\phi + \frac{1}{2}\heavi(-s_i^\kappa) x_i^\mu + \mathcal{N}\Big( 0, \frac{N_Z + 3N_E -2}{2N} \Big) \\
    & \approx \frac{1}{2} \Bigg[ \heavi(s_i^\kappa) x_i^\phi + \heavi(-s_i^\kappa) x_i^\mu + \sqrt{2} \cdot \mathcal{N}\Big( 0,\frac{N_Z + 3N_E}{N} \Big) \Bigg] \\
\end{split}
\label{eqn:apx_3}
\end{equation}
% \end{strip}
%
where in the third line we have made the same approximations as previously discussed. The postsynaptic sum is thus approximately $\vec{x}^\phi$ in all indices that are not currently being masked, which drives the network towards that (target) attractor. In vector form, the above is written as
\begin{equation}
    \mat{W} \big( \vec{x}^\mu \had \heavi ( \vec{s}^\kappa ) \big) \,\, \appropto \,\, \heavi ( \vec{s}^ \kappa) \had  \vec{x}^\phi + \heavi (- \vec{s}^ \kappa) \had  \vec{x}^\mu + \sqrt{2}\sigma\vec{n}
\end{equation}
where it is assumed that there exists a stored transition from state $\vec{x}^\mu$ to $\vec{x}^\phi$ with stimulus $\vec{s}^\kappa$, and $\appropto$ denotes approximate proportionality. A similar calculation can be performed in the case that a stimulus is imposed which does not correspond to a valid transition for the current state. In this case, no terms of significant magnitude emerge from the transition summation, and we are left with 
\begin{equation}
    \mat{W} \big( \vec{x}^\mu \had \heavi ( \vec{s}^\text{invalid} ) \big) \,\, \appropto \,\, \vec{x}^\mu + \sqrt{2}\sigma\vec{n}
\end{equation}
i.e. the attractor dynamics are largely unaffected. Since we have not distinguished between our above attractor terms being node attractors or edge attractors, or our stimuli from being $\vec{s}_a$ or $\vec{s}_b$ stimuli, the above results can be applied to all relevant situations \textit{mutatis mutandis}.

\subsection{Why model input as masking?}
\label{subsec:hadamard_input}
\begin{figure*}
  \centering
  % \vspace{-4em}
  % \begin{subfigure}{\linewidth}
  %   \centering
  %   \includegraphics[width=\linewidth]{figs/walk_sync_hadamard.pdf}
  % \end{subfigure}
  
  % \begin{subfigure}{\linewidth}
  %   \centering
  %   \includegraphics[width=\linewidth]{figs/walk_async_hadamard.pdf}
  % \end{subfigure}\
  \includegraphics[width=0.8\linewidth]{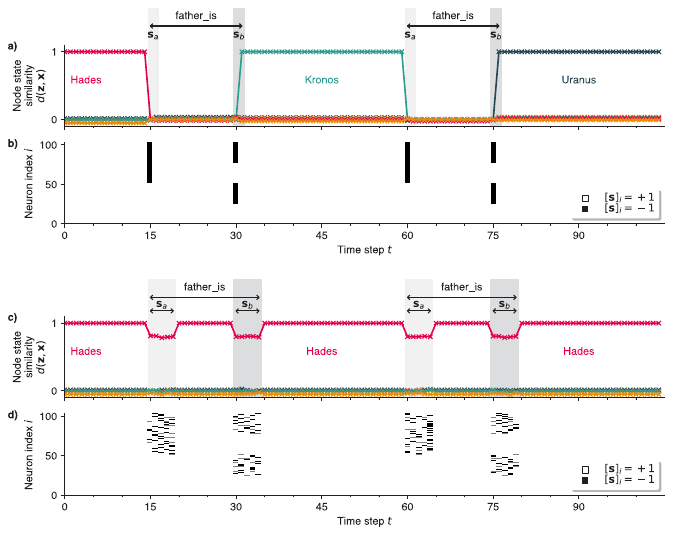}
  % \vspace{-4em}
  \caption{An attractor network constructed via the simpler weights construction method specified in Section \ref{subsec:hadamard_input}, with input to the network modelled as Hadamard product binding, rather than component-wise masking. \textbf{a)} The similarity of the network state $\vec{z}_t$ to stored node hypervectors, when the stimulus hypervector $\vec{s}$ is applied on one time step for all neurons \textit{simultaneously}. \textbf{b)} A subset of the stimulus hypervector $\vec{s}$ at each time step in this \textit{synchronous} case. \textbf{c)} The attractor overlaps in the \textit{asynchronous} case, where the stimulus $\vec{s}$ is applied over multiple time steps randomly. \textbf{d)} A subset of the stimulus hypervector $\vec{s}$ at each time step in this \textit{asynchronous} case. For visual clarity, the two stimulus hypervectors shown were manually chosen rather than randomly generated. In the synchronous case, the network performs the correct walk between attractor states as intended. In the asynchronous case however, the stimuli fail to effect the desired transitions, since any changes in the network state caused by the input stimuli are short-lived, as they are quickly reversed on the next time step by the attractor network's pattern-correcting dynamics.
  }
  \label{fig:hadamard_input}
\end{figure*}

One immediate question might be why we have chosen to model input to the network as a masking of the neural state vector \mbox{(Equation \ref{eqn:update_hop_mask})}, rather than simply modelling input as a Hadamard product, with a state update rule given by
\begin{equation}
\label{eqn:activation_hadamard}
    \vec{z}_{t+1} = \mathrm{sgn} \big( \mat{W} (\vec{z}_{t} \had \vec{s}) \big)
\end{equation}
such that a component for which the input stimulus $s_i = -1$ triggers a ``flip'' in the neuron state $+1 \leftrightarrow -1$. As will be shown, the problem with this construction is that it relies on the synchrony of input to the network, and does not allow for for the input to arrive asynchronously and with arbitrary delays. While this would not be a problem for a digital synchronous system, such timing constraints cannot be expected to be met in a network of asynchronously-firing biological neurons. In the synchronous case however, the edge terms $\mat{E}^{\eta}$ in the weights matrix construction could be simplified to
\begin{equation}
\mat{E}^{(\eta)} = \vec{y} \big( \vec{x} \had \vec{s} \big)^\intercal
\end{equation}
where as per previous notation, $\vec{x}$ and $\vec{y}$ are the source and target attractor states respectively, and $\vec{s}$ the stimulus to cause the transition. Superficially, this construction would then satisfy our main requirements for achieving the desired attractor itinerancy dynamics during input and rest scenarios, namely
\begin{equation}
    \mat{W} \vec{x} \approx \vec{x} \quad \forall \quad  \vec{x} \in X_{\mathrm{AN}}
\label{eqn:had_mult_attr}
\end{equation}
which ensures that while there is no input to the network, the states $\vec{x}$ are stable attractors of the network dynamics, and
\begin{equation}
    \mat{W} \big( \vec{x} \had \vec{s} \big) \approx \vec{y}
\label{eqn:had_mult_tran}
\end{equation}
which ensures that inputting the stimulus $\vec{s}$ triggers the desired transition. The resulting dynamics for this network - when input is entirely synchronous - are shown in \mbox{Figure \ref{fig:hadamard_input}}a , and indeed the network performs the desired walk.

We then test the functionality of the attractor network with Hadamard input when the exact simultaneous arrival of input stimuli cannot be guaranteed, i.e. the input to the network is asynchronous. To model this, we consider that the arrival time of the stimulus is component-wise randomly and uniformly spread over 5 time steps, rather than just one. The same attractor network receiving the same sequence of Hadamard-product stimuli, but now asynchronously, is shown in \mbox{Figure \ref{fig:hadamard_input}} c). The network does not perform the correct walk between attractor states, and instead remains localised near the initial attractor state across all time steps. This is due to the fact that, although when input is applied, the network begins to move away from the initial attractor state, these changes are immediately undone by the network's inherent attractor dynamics, since the neural state is still within the initial attractor's basin of attraction. Only when the timescale of the input is far faster than the timescale of the attractor dynamics (e.g. input is synchronous) may the input accumulate fast enough to escape the initial basin of attraction.

\begin{figure*}
    \centering
    \includegraphics[width=0.8\linewidth]{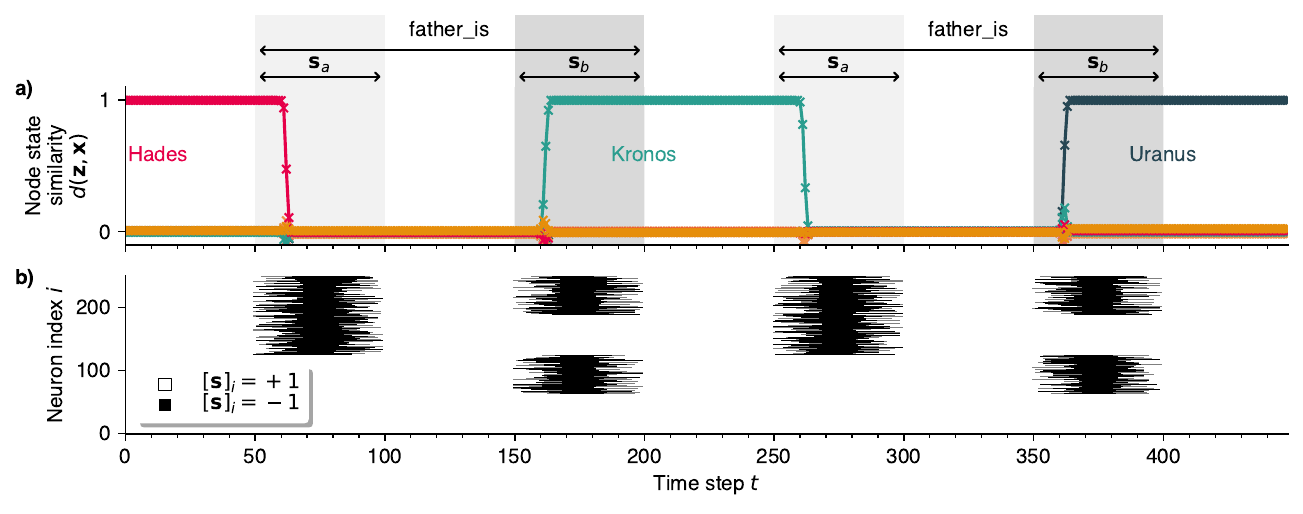}
    \caption{The attractor network performing a walk as masking input is applied asynchronously over multiple time steps with random delays. \textbf{a)} The similarities between the network state $\vec{z}_t$ and stored node hypervectors $\vec{x} \in X_{\mathrm{AN}}$. \textbf{b)} A subset of the stimulus hypervector $\vec{s}$ being applied to the network as a mask at each time step. Indices which are black on any time step have $[\vec{s}]_i = -1$ and so are being masked by the stimulus. For visual clarity, the two stimulus hypervectors shown were manually chosen, rather than randomly generated. The attractor transition dynamics are thus robust to input asynchrony when the input is modelled as a component-wise masking of the network state.}
    \label{fig:mask_inp_async}
\end{figure*}

When input to the network is treated as masking operation however \mbox{(Equation \ref{eqn:update_hop_mask})}, the attractor itinerancy dynamics are robust to input asynchrony. To model this, the input stimulus is stochastically applied, with each component being delayed randomly and uniformly by up to 20 time steps. The stimulus is then held for 10 time steps, and stochastically removed over 20 time steps in the same manner. The attractor network with asynchronous masking input is shown in \mbox{Figure \ref{fig:mask_inp_async}}, and functions as desired, performing the correct walk between attractor states. Modelling input to the network as a masking operation thus allows the network to operate robustly in asynchronous regimes, while modelling input to the network as a Hadamard product does not.

% ---------------------------------------------------------
\subsection{The need for edge states}
\label{sec:why_edge_states}

\begin{figure*}
  \centering
  % \vspace{-4em}
  % \begin{subfigure}{\linewidth}
  %   \centering
  %   \includegraphics[width=\linewidth]{figs/no_edge_states.pdf}
  % \end{subfigure}
  
  % \begin{subfigure}{\linewidth}
  %   \centering
  %   \includegraphics[width=\linewidth]{figs/with_edge_states_comparison.pdf}
  % \end{subfigure}
  \includegraphics[width=0.8\linewidth]{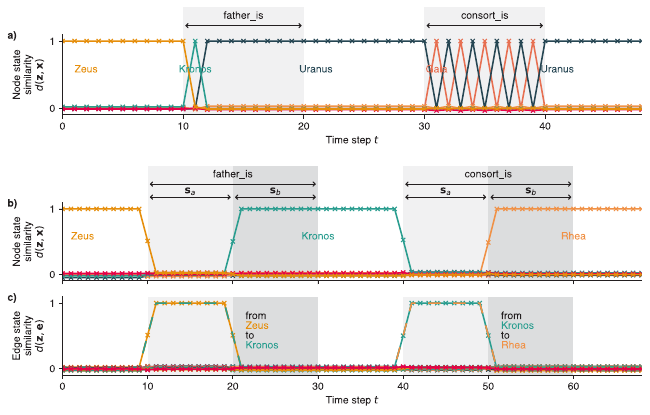}
  \caption{An attractor network receiving a sequence of stimuli to trigger a certain walk constructed \textbf{a)} without edge states and \textbf{b)} with edge states, with edge state overlaps being shown in \textbf{c)}. Due to the consecutive edges in the FSM \mbox{(Figure \ref{fig:starwars_net})} with the same stimulus ``father\_is'', the edge-state-less network overshoots and skips the ``Kronos'' state, stopping instead at the ``Uranus'' state. Similarly, there is an unwanted oscillation between the states ``Gaia'' and ``Uranus'' due to the bidirectional edge with stimulus ``consort\_is''. The addition of the edge state attractors resolves these issues, and allows the network to function robustly when input stimuli are applied for an arbitrary number of time steps.}
  \label{fig:why_edge_states}
\end{figure*}

The need for the edge state attractors arises when one wants to emulate an FSM where there are consecutive edges with the same stimulus. For example, in the FSM implemented throughout this article \mbox{(Figure \ref{fig:starwars_net})} there is an incoming edge from ``Zeus'' to ``Kronos'' with stimulus ``father\_is'' and then immediately an outgoing edge from ``Kronos'' to ``Uranus'' with stimulus ``father\_is'' also. More generally, consider that we wish to embed the transitions
\begin{equation}
\vec{x}_1 \stackrel{\vec{s}}{\longrightarrow} \vec{x}_2 \stackrel{\vec{s}}{\longrightarrow} \vec{x}_3 \end{equation}

In the fully synchronous case, i.e. when input is applied for one time step only, there is no need for edge states. When the stimulus $\vec{s}$ is applied, the network will make one transition only. In the asynchronous case however, one cannot ensure that the stimulus is applied for one time step only. Thus, starting from $\vec{x}_1$, when the stimulus is applied ``once'' for an arbitrary number of time steps, the network may have the unwanted behaviour of transitioning to $\vec{x}_2$ on the first time step, and then to $\vec{x}_3$ on the second, effectively overshooting and skipping $\vec{x}_2$. In \mbox{Figure \ref{fig:why_edge_states}} we see the dynamics of the attractor network constructed without any edge states, with inputs which are applied for 10 time steps each, and we indeed see the undesirable skipping behaviour. Similarly, bidirectional edges with the same stimulus (e.g. ``consort\_is'') cause an unwanted oscillation between attractor states. The edge states offer a solution to this problem: by adding an intermediate attractor state for every edge, and splitting each edge into two transitions with stimuli $\vec{s}_a$ and $\vec{s}_b$, we ensure that there are no consecutive edges with the same stimulus.

\begin{figure*}
\centering
% \vspace{-4em}
% \begin{subfigure}{\linewidth}
% \centering
% \includegraphics[width=3in]{figs/carrot_graph.pdf}
% \vspace{1em}
% \end{subfigure}

% \centering
% \begin{subfigure}{\linewidth}
% \centering
% \includegraphics[width=5in]{figs/walk_pointy_carrot.pdf}
% \end{subfigure}

% \begin{subfigure}{\linewidth}
% \centering
% \includegraphics[width=5in]{figs/walk_round_clementine.pdf}
% \end{subfigure}
\includegraphics[width=5in]{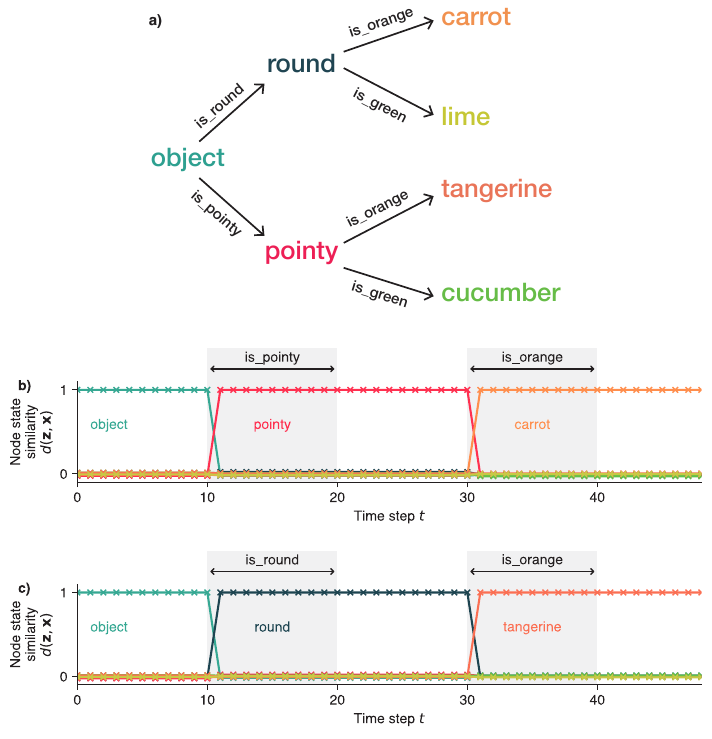}

\caption{
Embedding an FSM that does not require edge states, since it does not have consecutive edges with the same stimulus. \textbf{a)} The FSM to be embedded, representing a simple decision tree. \textbf{b)} \& \textbf{c)} An attractor network constructed to store this FSM, without any edge states, as a sequence of stimuli is input. The network performs the correct walks between attractor states as desired. To note is that the second stimulus (``is\_orange'') and its transition are state-dependent, as the target state (``carrot'' or ``tangerine'') is dependent upon the stimulus given 20 time steps before (``is\_round'' or ``is\_pointy''). This illustrates that the edge states are not strictly necessary to implement state-dependent transitions between attractor states.}
\label{fig:pointy_carrot}
\end{figure*}

If we don't necessarily need to be able to embed FSMs with consecutive edges with the same stimulus, then we can rid of the edge states, and construct our weights matrix with simpler transition terms like in Equation \ref{eqn:dw_no_edge}. An attractor network constructed in this way is shown in \mbox{Figure \ref{fig:pointy_carrot}}, for a chosen FSM that does not require edge states, but still contains state-dependent transitions. The network performs the correct walk between attractor states as intended, and does not suffer from any of the unwanted skipping or oscillatory phenomena like in \mbox{Figure \ref{fig:why_edge_states}}. Thus, while the edge states are required to ensure that any FSM can be implemented in a ``large enough'' attractor network, they are not strictly necessary to achieve state-dependent stimulus-triggered attractor transition dynamics.

\subsection{Sparse stimuli}
\label{section:sparse_stimuli}

\begin{figure*}
\centering
\includegraphics[width=0.9\linewidth]{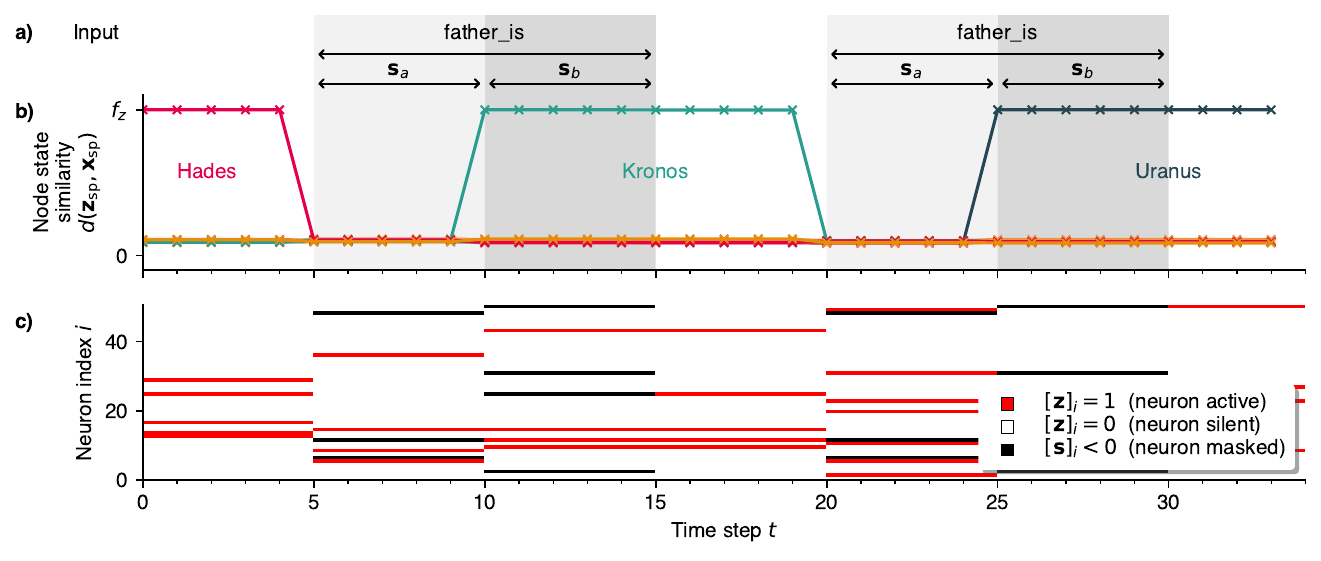}
\caption{An attractor network with both sparse states and sparse stimuli, constructed as described in Section \ref{section:sparse_stimuli}. The values used here are $N = 10,000$, $f_z = 0.1$ (meaning only 10\% of neurons are active at any time), and $f_s = 0.9$ (meaning that only 10\% of neurons are masked by the stimulus). \textbf{a)} The input hypervectors to the network, each masking a random 10\% of neurons within the network. \textbf{b)} The overlap of the sparse network state $\vec{z}_\text{sp}$ with stored attractor hypervectors. \textbf{c)} A subset of the neurons within the network, showing the active neurons ($z_j = 1$) in red, as well as which neurons are currently being masked by the input ($s_j < 0$). The network performs the correct walk between attractor states. The balanced bipolar stimulus hypervectors used throughout this paper may thus also be generalised to be sparse.}
\label{fig:sparse_stim}
\end{figure*}

One shortcoming of the model might be that we used dense bipolar hypervectors $\vec{s}$ to represent the stimuli, meaning that when $\vec{s}$ is being input to the network, masking all neurons for which $s_j = -1$, approximately half of all neurons within the network are silenced. This was initially chosen because unbiased bipolar hypervectors are arguably the simplest and most common choice of VSA representation, and highlights the fact that VSA-based methods can be applied to the design of attractor networks with very little required tweaking \cite{gayler_multiplicative_1998, kleyko_survey_2022}.

From the biological perspective however, it could be seen as somewhat implausible that the number of active neurons should change so drastically (halving) while a stimulus is present. Furthermore, if implemented with spiking neurons, the large changes in the total spiking activity could cause unwanted effects in the spike rate of the non-masked neurons. Also, this means that while the network is being masked, the size of the network (and so its capacity) is reduced to $N/2$, and so the network is especially prone to instability during the transition periods, if the network is nearing its memory capacity limits.

For these reasons, it is worth exploring whether the network could be constructed such that during a masking operation, fewer than half of all neurons are masked, i.e. $\vec{s}$ is biased to contain more $+1$ than $-1$ entries\footnote{We could also use binary $\vec{s}$ hypervectors, rather than positive/negative, and then alter the transition terms $\mat{E}^\eta$ to include $f$ and $1/(1-f)$ terms to achieve the same result. We believe it is more intuitive not to make this change for this section, however.}. To keep the notation consistent with the notation used for sparse binary hypervectors, we will denote the coding level of the attractor states as $f_z$ (where previously it was simply $f$) and the coding level of the stimulus hypervectors as $f_s$. The coding level of the stimulus hypervectors $f_s$ we define to be the fraction of components for which $s_j > 0$. A stimulus hypervector with $f_s > 0.5$ thus silences fewer neurons from the network during a masking operation.
This is not the only change we need to make however. If we turn to our (sparse) edge terms \mbox{(Equation \ref{eqn:tran_term_sparse})}, they were previously constructed such that they would produce a non-negligible overlap with the network state $\vec{z}_\text{sp}$ if and only if the network is in the correct attractor state \textit{and} is being masked by the correct stimulus. The important condition to be fulfilled is then
\begin{equation}
\begin{split}
    \mathbb{E} \Big[ \big[ ( (\vec{x}_\text{sp}-f\vec{1})\had \vec{s} )\cdot \vec{x}_\text{sp} \big]_j \Big] \stackrel{!}{=} 0 \quad \forall \quad j = 1\ldots N
\end{split}
\end{equation}
% and
% \begin{equation}
% \begin{split}
%     \mathbb{E} \Big[ \big[ ( (\vec{x}_\text{sp}-f\vec{1})\had \vec{s} )\cdot \vec{y}_\text{sp} \big]_j \Big] \stackrel{!}{=} 0 \quad \forall \quad j = 1\ldots N
% \end{split}
% \end{equation}
% %

that is, the overlap should be negligible if the network is in the correct attractor state, but the stimulus is \textit{not} present. This condition is satisfied if the components of $\vec{s}$ are generated according to
%
% \FloatBarrier
\begin{table}[h!]
\centering
\begin{tabular}{l|l}
$s$ & $\probP (s_j=s)$ \\ \hline
$1$  &   $f$     \\
$-f / (1-f)$  & $(1-f)$      
\end{tabular}
\end{table}
% \FloatBarrier
%
\noindent 
\noindent where $s_j$ is the $j$'th component of $\vec{s}$. This implies that for a stimulus hypervector biased towards having more positive entries (fewer neurons are masked), the negative entries must increase in magnitude to compensate for their infrequency. For the case that only a quarter of neurons are masked by the stimulus ($f_s = 0.75$), the negative 25\% of components must have the value $-3$, while for $f_s = 0.5$ this of course collapses to the balanced bipolar hypervectors used throughout this article with $\probP(S_j = 1) = \probP(S_j = -1) = 0.5$ (Equation \ref{eqn:probX_half}). We are forced to increase the magnitude of the negative terms, rather than reduce the positive terms, since the magnitude of the positive terms must remains identical to that of the stored attractor terms, in order to ensure that the correct target state is projected out during a transition. We can then construct our weights matrix in the same way as before, but using these biased stimulus hypervectors $\vec{s}$. An attractor network was generated with coding levels $f_z = 0.1$ (10\% of neurons are active in any attractor hypervector) and $f_s = 0.9$ (10\% of neurons are masked by stimulus hypervectors), and the results are shown in \mbox{Figure \ref{fig:sparse_stim}}, with the neural state performing the correct walk between attractor states as desired.

To be noted is that as we approach $f_s \rightarrow 1$, the stimuli become less and less distributed, with the limiting case $f_s = 1 - 1/N$ implying that only one component of $\vec{s}$ is negative, and so by masking only one neuron, the network will switch between attractor states. This case is obviously a stark departure from the robustness which the more distributed representations afford us, since if that single neuron is faulty or dies, it would be catastrophic for the functioning of the network. Similarly, if another independent stimulus were to, by chance, choose the same component to be non-negative, this would cause similarly unwanted dynamics. Less catastrophic, but still worth considering is that the noise added per edge term, as a result of the negative terms becoming very large, has variance that scales like $\mathrm{Var}[s_j] \approx 1/(1-f_s)$, and so for $f_s \rightarrow 1$ contributes an increasing amount of unwanted noise to the system, destabilising the attractor dynamics. Nevertheless, this represents yet another trade-off in the attractor network's design, as needing to mask fewer neurons might be worth the increased noise within the system, decreasing its memory capacity.

\FloatBarrier

\begin{table*}
\centering
\begin{tabular}{ll}
Symbol & Definition \\ \hline
$N$ & Number of neurons within the attractor network \\
$N_Z$ & Number of FSM states\\
$N_E$ & Number of FSM edges\\
$\vec{a}, \vec{b}, \vec{c} \ldots$      &    Dense bipolar hypervectors       \\
$\vec{a}_{\mathrm{sp}}, \vec{b}_{\mathrm{sp}}, \vec{c}_{\mathrm{sp}} \ldots$      &    Sparse binary hypervectors       \\
$f$ & Coding level of a hypervector (fraction nonzero components) \\
$\vec{z}_t $ &  Neuron state vector at time step $t$ \\
$\vec{x}, \vec{y}$  & Node hypervectors representing an FSM state \\
$\vec{e}$       & Edge-state hypervectors \\
$\vec{s}, \vec{s}_a, \vec{s}_b$      & Stimulus hypervectors \\
$\vec{r}$      & Ternary output hypervectors \\
$\vec{1}$ & A hypervector of all ones \\
$\mat{W}$      & Recurrent weights matrix\\
$w_{ij}$      & Synaptic weight from neuron $j$ to $i$ \\
$\mat{E}$      & Matrices added to $\mat{W}$ to implement transitions \\
$\had$ & Hadamard product (component-wise multiplication) \\
$\vec{x}\T$ & Transpose of $\vec{x}$ \\
$\heavi (\cdot )$ & Component-wise Heaviside function \\
$\mathrm{sgn} (\cdot )$ & Component-wise sign function \\
\end{tabular}
\caption{Notation and frequently used symbols.}
\end{table*}

\FloatBarrier

% \nolinenumbers

\end{document}